\documentclass[journal]{IEEEtran}
\IEEEoverridecommandlockouts
\usepackage{cite}
\usepackage{amsmath,amssymb,amsfonts}
\usepackage{subcaption}
\usepackage[normalem]{ulem}
\usepackage{tikz}
\usepackage{tabularx}
\usepackage{graphics}
\usepackage{adjustbox}
\usepackage{enumerate,graphics}
\usepackage[english]{babel}
\usepackage{array}
\usepackage{multirow}
\usepackage{algorithm}
\usepackage{algorithmic}
\usepackage{times}
\usepackage{amsmath}
\usepackage{amssymb}
\usepackage{enumitem}
\usepackage[dvips]{epsfig}
\usepackage{epstopdf}
\newtheorem{Def}{\textbf{Definition}}

\begin{document}

\title{An Improved Transfer Model: Randomized Transferable Machine}

\author{Pengfei Wei$^{1}$,
        Xinghua Qu$^{2}$, Yew Soon Ong$^{2,3}$, and Zejun Ma$^{1}$ \\
 $^1$AI Lab, Bytedance, Singapore.  \\
 $^2$School of Computer Science and Engineering, Nanyang Technological University, Singapore. \\
 $^3$Agency for Science, Technology and Research, Singapore (A*STAR). \\
}

\maketitle

\begin{abstract}
Feature-based transfer is one of the most effective methodologies for transfer learning.
Existing studies usually assume that the learned new feature representation is \emph{domain-invariant}, and thus train a transfer model $\mathcal{M}$ on the source domain.
In this paper, we consider a more realistic scenario where the new feature representation is suboptimal and small divergence still exists across domains.
We propose a new transfer model called Randomized Transferable Machine (RTM) to handle such small divergence of domains.
Specifically, we work on the new source and target data learned from existing feature-based transfer methods.
The key idea is to enlarge source training data populations by randomly corrupting the new source data using some noises, and then train a transfer model $\widetilde{\mathcal{M}}$ that performs well on all the corrupted source data populations.
In principle, the more corruptions are made, the higher the probability of the new target data can be covered by the constructed source data populations, and thus better transfer performance can be achieved by $\widetilde{\mathcal{M}}$.
An ideal case is with infinite corruptions, which however is infeasible in reality.
We develop a marginalized solution that enables to train an $\widetilde{\mathcal{M}}$ without conducting any corruption but equivalent to be trained using infinite source noisy data populations.
We further propose two instantiations of $\widetilde{\mathcal{M}}$, which theoretically show the transfer superiority over the conventional transfer model $\mathcal{M}$.
More importantly, both instantiations have closed-form solutions, leading to a fast and efficient training process.
Experiments on various real-world transfer tasks show that RTM is a promising transfer model.
\end{abstract}


\section{Introduction}
Transfer learning \cite{tan2018survey} aims to build a transfer model using the data from a source domain while generalizing good performance on the prediction task of another different but related target domain.
Such a knowledge transfer avoids the expensive process of collecting target labelled data, but faces the challenge of domain divergence, i.e., the data distribution usually differs dramatically across domains.
Feature-based transfer method is one of the most effective methodologies to handle the distribution discrepancy, where subspace learning \cite{wang2019softly,li2019multisource,wei2021subdomain} and deep feature learning \cite{Wei2016DeepNF,Long2017Deep,zhang2018collaborative} are two popular research lines.
The underlying rationale is that, although two domains may differ considerably in the original feature space, there exists a \emph{domain-invariant feature representation with which the data distribution discrepancy can be eliminated}.
Both subspace learning and deep feature learning methods aim to learn such a `good' feature representation.

To yield better and better feature representation, researchers are exploring more and more complex data properties, including geometric structure \cite{pan2011domain}, sparsity \cite{long2013TSC}, second-order moments \cite{sun2016return}, low-rankness \cite{Li2018Domain}, multiple manifolds \cite{wei2019knowledge}, and disentanglement \cite{cai2019learning}, etc.
With the recent advancement in deep learning, research efforts are gradually confined to the improvements on the network structure.
With the learned new feature representation, existing works directly train a transfer model on the source domain.
Situation differs slightly between subspace learning and deep feature learning, where the former usually decouples feature learning with model training as two separate stages while the latter jointly learns new feature representation and transfer model together.
Admittedly, such a conventional way of training transfer model is straightforward considering that only the source domain has sufficient labelled data.
However, an important question remains: Is the learned new feature representation sufficiently good enough to guarantee that the transfer model trained on the source domain can generalize well to the target domain?
In other words, is the new feature representation truly \emph{domain-invariant}?
\begin{figure*}[t]
\centering
 \subfloat[ Original data ]
{\includegraphics[scale = 0.08]{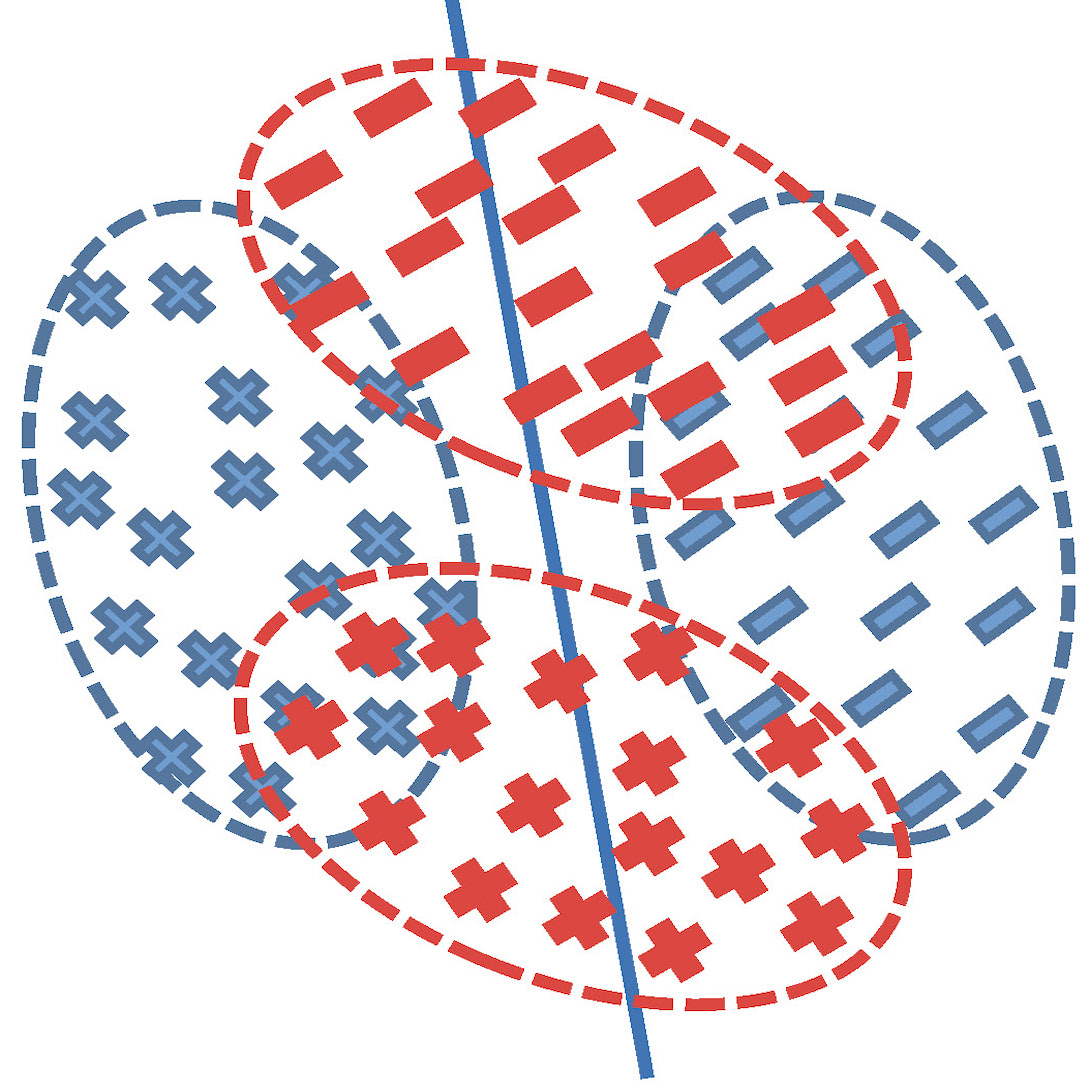} \label{shownexample1}} \ \ \ \ \ \ \
\subfloat[ Transfer model on new data ]
{\includegraphics[scale = 0.7]{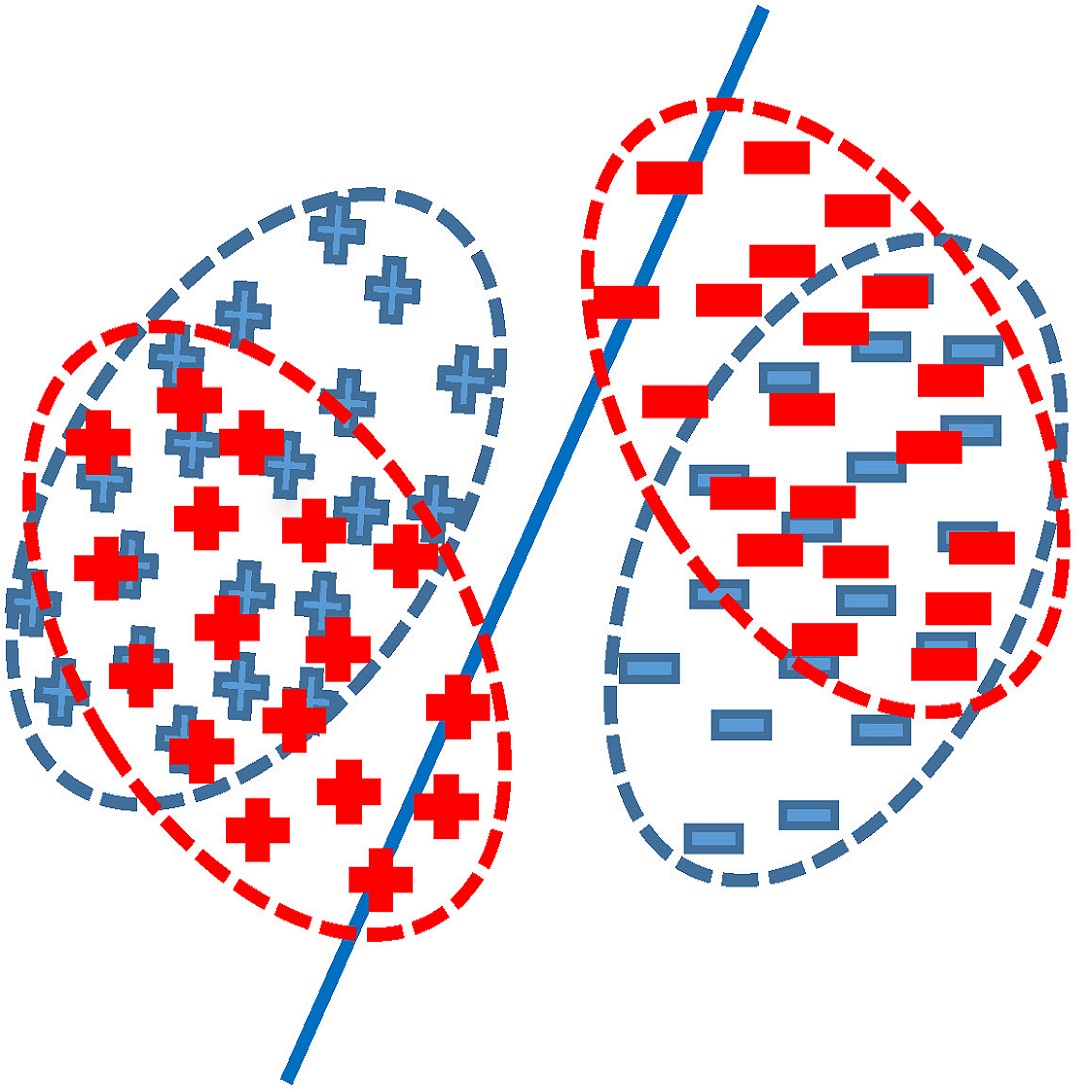} \label{shownexample2}} \ \ \ \ \ \ \
\subfloat[ Noisy source data ]
{\includegraphics[scale = 0.7]{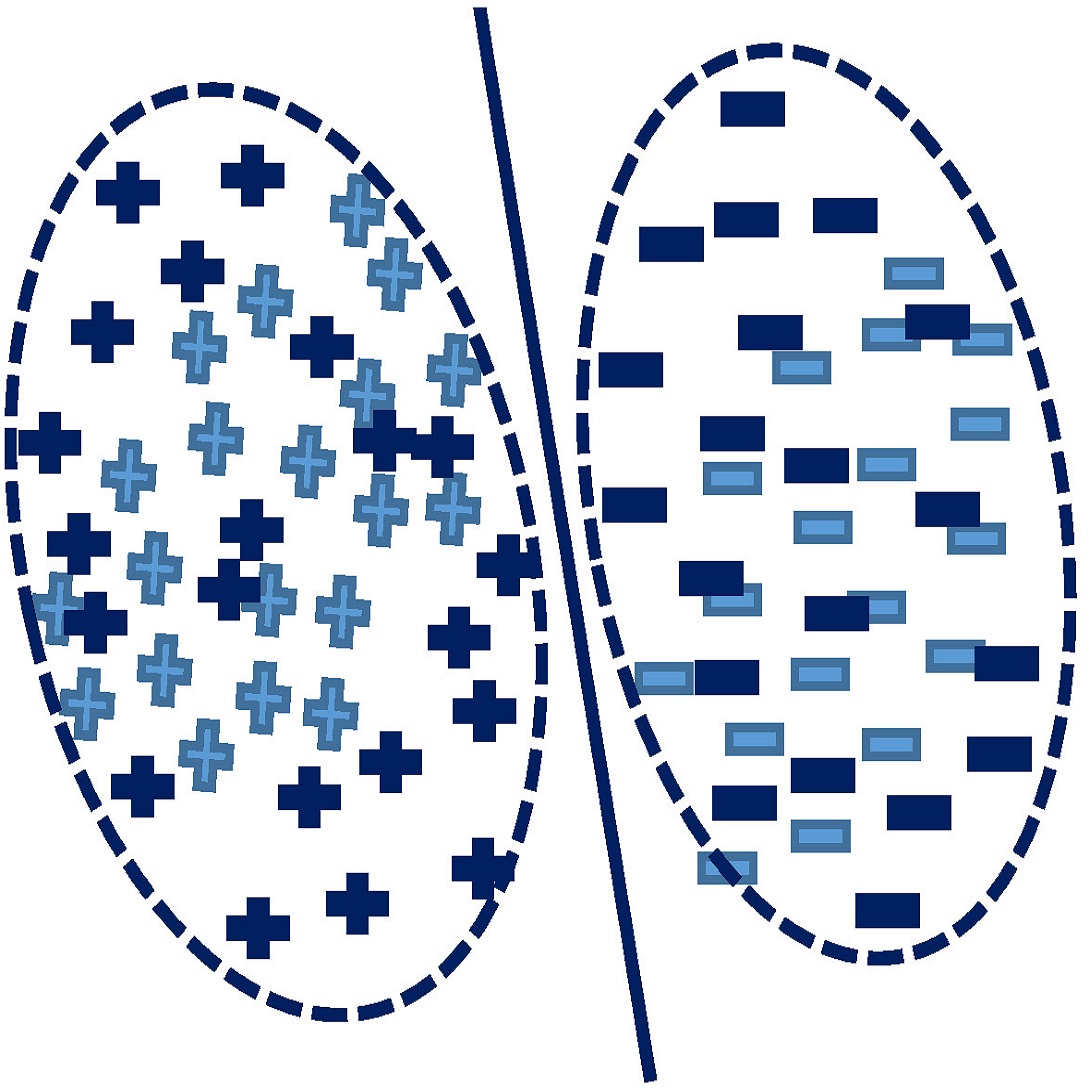} \label{shownexample3}} \ \ \ \ \ \  \
 \subfloat[New transfer model]
{\includegraphics[scale = 0.7]{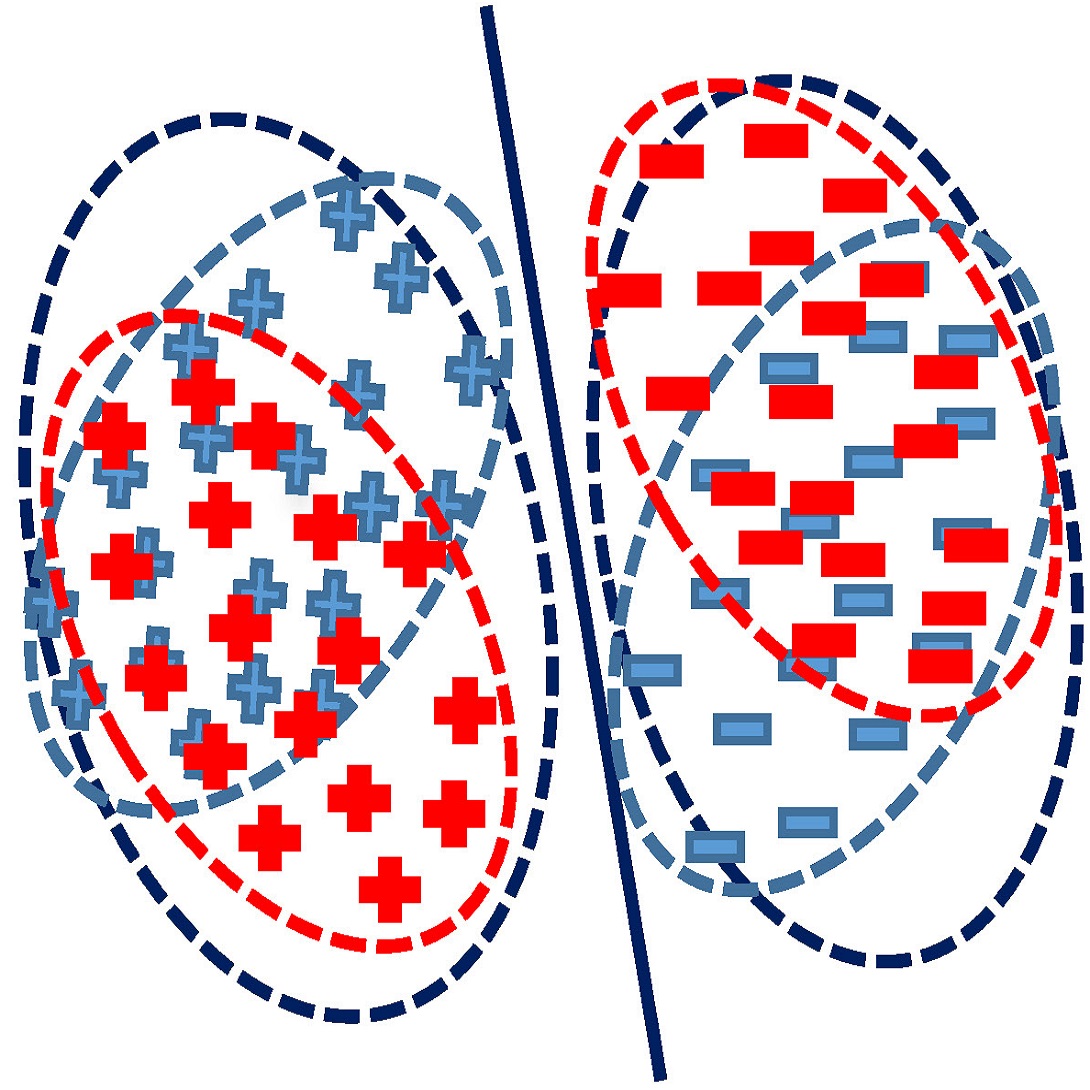} \label{shownexample4}}
 \caption{The red are the target data, the light blue are the original source data, and the dark blue are the constructed noisy source data. The `+'s are positive samples, the `-'s are negative samples, and the solid line is the classification hyperplane. 
 } \label{shownexample}
\end{figure*}

The continuing emergence of new feature-based transfer methods with their incremental improvements on transfer performance implicitly indicates that the truly \emph{domain-invariant} feature representation may have not been achieved yet.
Although existing efforts claim the success of learning the \emph{domain-invariant} feature representation, their results are usually quickly ``defeated" by newer works.
New feature-based transfer methods are flourishing, and such a `defeat-to-improve' situation seems to be endless as it stands now.
One may naturally doubt whether the truly \emph{domain-invariant} feature representation is achievable.
Unfortunately, no existing works can answer this question, and we can only get some clues from the development of feature-based transfer studies.

The distance metric, used to measure the feature distribution discrepancy of domains, is diversified including Maximum Mean Discrepancy (MMD) \cite{borgwardt2006integrating}, A-distance \cite{ganin2016domain}, and Wasserstein distance (W-distance) \cite{lee2019sliced}, etc.
There is no a clear winner that is significantly superior over the others among these metrics.
In practice, it is difficult to decide which metric is the best for transfer learning, and infeasible to predict whether more advanced metrics would be proposed in the future.
Moreover, for a specific metric, it usually cannot achieve the ideal zero-divergence due to various reasons, e.g., the finite number of data samples, the consideration of other data properties, the hyper-parameters initialized for learning, and optimization related matters.
With these observations, we deduce that it is extremely hard to achieve the truly \emph{domain-invariant} feature representation.
Current works may only yield some suboptimal alternatives.

Consequently, there is still space to further boost the transfer performance, although \emph{the data with the new feature representation} (called the new data in the following sections) have enhanced domain similarities.
Since distribution discrepancy still exists in the new data of the domains, probably very small, \emph{the conventional transfer model directly trained on the new source data} is problematic.
We thus focus on the problem of how to train an improved transfer model that further boosts the transfer performance of the conventional one.
To the best of our knowledge, this is the first work studying such a problem, while existing works assume the new data are truly \emph{domain-invariant}, and directly adopts the conventional transfer model.

Considering that the new data of the source and target domains are already similar in many extents, we should pay more attention to the nuances of the domains in the learning of the improved transfer model.
To do so, we propose randomized transferable machine (RTM).
Instead of training a model that fits the new source data only, we propose to train an RTM that is capable of generalizing good performance on numerous noisy data populations that are generated from the new source data.
The underlying intuition is that the noisy data populations are diversified enough to cover the characteristics of the new target data.
This is reasonable as the new source and target data are already very similar.
The small discrepancies can be easily captured by the rich noisy source data populations.
Figure \ref{shownexample} shows the idea of RTM.
From Figure \ref{shownexample1} to \ref{shownexample2}, we can see that the two domains represented by the new data are well aligned.
Only a small region of the new target data is wrongly classified by the transfer model trained on the source.
To further make correctness on this small region, a large number of noisy data are generated by randomly corrupting the new source data as shown in Figure \ref{shownexample3}.
A hyperplane $\mathcal{H}$ is then learned on these noisy source data populations.
Figure \ref{shownexample4} shows that the constructed noisy source data populations successfully cover the new target data, and $\mathcal{H}$ performs well.

In principle, the more random corruptions are constructed, the higher the probability of the new target data can be covered by the noisy source data populations, and thus the better transfer performance is achieved by RTM.
The ideal case is training RTM with an infinite number of random corruptions.
However, it is infeasible in practice.
To make it happen, we provide a marginalized solution to simulate infinite random corruptions without really conducting any corruption.
By training RTM with this marginalization trick, it is expected to boost transfer performance on the target task.
Note that different types of noise can be used to corrupt the new source data.
In this paper, we present the general form of RTM, and then propose two instantiations of RTM with a dropout noise.
By using the linear regression model and the kernel ridge regression model as the base model, we develop an RTM$_{lr}^{d}$ and an RTM$_{kr}^{d}$.
The learning objectives of the two instantiations have closed-form solutions, which makes the training of them fast and efficient.
Since RTM$_{lr}^{d}$ and RTM$_{kr}^{d}$ have the similar mathematical form, we only focus on the easier one, i.e., RTM$_{lr}^{d}$, to show the effectiveness of the idea of RTM.
We test RTM$_{lr}^{d}$ using the new data learned from various state-of-the-art feature-based transfer methods.
The empirical results show that RTM$_{lr}^{d}$ further improves the transfer performance compared with the conventional transfer model.
We also test RTM$_{lr}^{d}$ by directly training it using the original source data, and surprisingly, we observe that RTM$_{lr}^{d}$ is capable of achieving considerable good performance.
In some tasks, RTM$_{lr}^{d}$ even outperforms deep learning based transfer methods.
Further considering the fast training speed, we believe RTM$_{lr}^{d}$ is a promising transfer model.

This paper is an extension of \cite{wei2021randomized}. Compared with \cite{wei2021randomized}, we make the following new contributions in this paper. 
\begin{itemize}
\item Firstly, we rewrote the introduction and related work sections to make them more comprehensive. We added more discussions on the state-of-the-art studies and highlighted the motivation of our method.
\item Secondly, we proposed a new general framework of Randomized Transferable Machine (RTM) that allows diversified instantiations. The paper \cite{wei2021randomized} is one of the instantiations with a linear regression function and a dropout noise.
We further proposed a new instantiation with a kernel ridge function and the dropout noise.
\item Thirdly, we gave more theoretical analyses on the underlying rationale why the instantiated RTM is superior to the conventional transfer model on the transfer performance from the perspective of adaptive regularization.
\item Fourthly, we presented a comprehensive discussion on the proposed RTM including the relatedness to other research topics, the generality to different instantiations, the training computational complexity, the technical significance compared with the existing works, the scalability to different feature-based transfer methods, and the applicability to the original data.
\item Lastly, we conducted extensive experimental studies to verify the transfer performance of RTM. The new experiments include a set of synthetic experiments, experiments on new feature representation learned from deep learning based methods, model similarity analysis, and data augmentation analysis.
\end{itemize}

The rest of the paper is organized as follows.
We discuss related work in Section II.
In Section III, we elaborate the problem setting and the motivation of our RTM.
We then introduce the technical details of RTM in Section IV and V.
The experimental study is described in Section VI.
We conclude the paper in Section VII.

\section{Related Work}
Feature-based transfer learning methods have been widely studied in the last decade.
The underlying intuition is that there exists a \emph{domain-invariant} feature representation across domains.
One typical way of learning \emph{domain-invariant} feature representation is by constructing subspace.
In \cite{pan2011domain}, Pan et al. propose to learn the transfer component of different domains by minimizing the marginal distribution discrepancies measured by MMD.
To model the domain shift dynamically, Gong et al. \cite{gong2012geodesic} propose to generate an
infinite number of intermediate subspaces that lie between the source and target domains along the geodesic flow on a Grassmann manifold.
In order to alleviate the computational issue of \cite{gong2012geodesic}, Fernando et al. develop a subspace alignment method \cite{fernando2013unsupervised} that directly aligns the subspaces of different domains.
Following studies consider diverse data properties in transfer.
Long et al. \cite{long2013transfer} not only align the marginal distributions of different domains, but also align the conditional ones by using pseudo target labels.
They further propose a transfer joint matching \cite{long2014transfer} method that jointly matches feature distributions and reweights instances.
Sun et al. \cite{sun2016return} consider the second-order statistics to align domains.
Aiming at benefiting from both the geometric property and the statistical property of data, Zhang et al. \cite{zhang2017joint} propose a joint geometrical and statistical alignment for visual domain adaptation.
Most recently, Wei et al. \cite{wei2019knowledge} propose to learn shared feature representations based on a multiple manifolds assumption.

With the development of deep neural networks, deep feature learning based transfer method is attracting increasing interests.
Chen et al. \cite{chen2012marginalized} propose a marginalized stacked denoising autoencoder to learn deep features by reconstructing the randomly corrupted data.
Deep nonlinear feature coding \cite{Wei2016DeepNF} further boosts the transfer performance of \cite{chen2012marginalized} by considering the explicit feature alignment and the nonlinearity exploration.
Long et al. \cite{long2015learning} develop a deep adaptation network (DAN) to learn transferable features.
Multiple kernelized MMD is adopted to reduce domain discrepancies in the last three layers of DAN.
Instead of focusing on the last three layers, a joint adaptation network (JDN) \cite{Long2017Deep} is proposed to minimize the joint distribution of full layers of features through MMD.
Considering the class weight bias across domains, Yan et al. \cite{Yan2017Mind} develop a weighted domain adaptation network (WDAN), which improves the transfer performance of DAN.
Apart from using MMD as distance metric, Ganin et al. \cite{ganin2016domain} propose an adversarial learning based method for transfer.
It builds a domain discriminator to confuse the task of classifying domains.
Following \cite{ganin2016domain}, Tzeng et al. \cite{tzeng2017adversarial} propose to combine discriminative modelling, weight sharing, and generative adversarial learning.
Cai et al. \cite{cai2019learning} explore semantic representations by the disentanglement.
A variational auto-encoder to reconstruct semantic latent variables and domain latent variables is proposed.
Although generally more effective than subspace-based methods, many deep transfer learning methods are actually motivated from subspace-based methods, e.g., DAN \cite{long2015learning} is motivated by TCA \cite{pan2011domain}.

\section{Problem Setting and Motivation}

\subsection{Problem Outline}
We focus on the knowledge transfer between two domains, one source domain $\mathcal{S}$ and one target domain $\mathcal{T}$.
Data from $\mathcal{S}$ and $\mathcal{T}$ lie in the same feature space but are with different feature distributions.
The source domain $\mathcal{S}$ has sufficient labelled data while the target domain $\mathcal{T}$ only has unlabelled data.
We denote the original source data matrix with its corresponding label matrix as $\mathbf{X}^{\mathcal{S}} = [\mathbf{x}_1^\mathcal{S},...,\mathbf{x}_n^\mathcal{S}] \in \mathbb{R}^{d \times n}$ and $\mathbf{Y}^{\mathcal{S}} = [\mathbf{y}_1^\mathcal{S},...,\mathbf{y}_n^\mathcal{S}] \in \mathbb{R}^{C \times n}$ where $d$ is the dimensionality of the original feature space and $C$ is the number of class.
Herein, $\mathbf{x}_i^\mathcal{S}$ is the $i$-\emph{th} source instance, and $\mathbf{y}_i^\mathcal{S}$ is its one-hot label vector.
Similarly, we denote the target data matrix as  $\mathbf{X}^{\mathcal{T}} = [\mathbf{x}_1^\mathcal{T},...,\mathbf{x}_m^\mathcal{T}] \in \mathbb{R}^{d \times m}$.
The notations used in this paper are summarized in Table \ref{notation}.
The objective is to use $\mathbf{X} = [\mathbf{X}^\mathcal{S},\mathbf{X}^\mathcal{T}]$ and $\mathbf{Y}^\mathcal{S}$ to train a model, and then utilize the model to predict the labels of $\mathbf{X}^\mathcal{T}$.
A simple and brute-force way to do so is to train a model using $\mathbf{X}^\mathcal{S}$ and $\mathbf{Y}^\mathcal{S}$, denoted as $\mathcal{M}_{x}$, and then use $\mathcal{M}_{x}$ for the prediction of $\mathbf{X}^\mathcal{T}$.
However, due to the distribution discrepancy of $\mathbf{X}^\mathcal{S}$ and $\mathbf{X}^\mathcal{T}$, $\mathcal{M}_{x}$ is not capable of generalizing good performance on $\mathbf{X}^\mathcal{T}$.
\begin{table}[t]
    \centering
    \small
    \caption{Notations and descriptions}
    \renewcommand\arraystretch{1}
    \label{notation}
    \resizebox{\linewidth}{!}{
       \begin{tabular}[t]{|c||l|}
       \hline
         Notation  &Description \\
       \hline
       \hline
       $\mathbf{X}^{\mathcal{S}} $, \ $\mathbf{X}^{\mathcal{T}}$   &original source and target data matrix \\
       \hline
       $\mathbf{Y}^{\mathcal{S}} $, $\mathbf{y}^{\mathcal{S}}$   &source label matrix and source one-hot label vector \\
       \hline
       $\mathbf{z}_i^\mathcal{S}$    &the $i$-\emph{th} source instance with new feature representation \\
       \hline
       $\widetilde{\mathbf{z}}_{i,j}^\mathcal{S}$    &the $j$-\emph{th} corrupted version of $\mathbf{z}_i^\mathcal{S}$  \\
       \hline
       $\mathbf{Z}^\mathcal{S}$    &source data matrix with new feature representation \\
       \hline
       $\widehat{\mathbf{Z}}^\mathcal{S}$    &the $J$-times repeated version of ${\mathbf{Z}}^\mathcal{S}$ \\
       \hline
        $\widetilde{\mathbf{Z}}^\mathcal{S}$    &the corrupted version of $\widetilde{\mathbf{Z}}^\mathcal{S}$ \\
       \hline
       $\widehat{\mathbf{Y}}^\mathcal{S}$    &the $J$-times repeated version of ${\mathbf{Y}}^\mathcal{S}$ \\
       \hline
       $\mathcal{M}_x$   &conventional transfer model trained on $\mathbf{X}^\mathcal{S}$ and $\mathbf{Y}^\mathcal{S}$  \\
       \hline
       $\mathcal{M}_z$   &conventional transfer model trained on $\mathbf{Z}^\mathcal{S}$ and $\mathbf{Y}^\mathcal{S}$  \\
       \hline
       $\widetilde{\mathcal{M}}_x$   &RTM trained on $\mathbf{X}^\mathcal{S}$ and $\mathbf{Y}^\mathcal{S}$  \\
       \hline
       $\widetilde{\mathcal{M}}_z$   &RTM trained on $\mathbf{Z}^\mathcal{S}$ and $\mathbf{Y}^\mathcal{S}$  \\
       \hline
       \end{tabular}}
\end{table}

Feature-based transfer methods including subspace learning and deep feature learning have shown be effective.
Using feature-based transfer methods, one can obtain the new data for both domains.
Specifically, the new data matrices for $\mathcal{S}$ and $\mathcal{T}$ are denoted as $\mathbf{Z}^{\mathcal{S}} = [\mathbf{z}_1^\mathcal{S},...,\mathbf{z}_n^\mathcal{S}] \in \mathbb{R}^{k \times n}$ and $\mathbf{Z}^{\mathcal{T}} = [\mathbf{z}_1^\mathcal{T},...,\mathbf{z}_m^\mathcal{T}] \in \mathbb{R}^{k \times m}$, respectively, where $k$ is the dimensionality of the new feature representation.
With the new data, existing methods train a transfer model $\mathcal{M}_{z}$ using $\mathbf{Z}^\mathcal{S}$ and $\mathbf{Y}^\mathcal{S}$.
The prediction is done by applying $\mathcal{M}_{z}$ to $\mathbf{Z}^\mathcal{T}$.
The superiority of ${\mathcal{M}}_{z}$ over $\mathcal{M}_{x}$ has been verified in many feature-based transfer methods on various real-world transfer tasks.

\subsection{Motivation}
While affirming the transfer effectiveness of ${\mathcal{M}}_{z}$, we notice a basic assumption of ${\mathcal{M}}_{z}$, that is, the new feature representation is truly \emph{domain-invariant}.
Only under this assumption, can $\mathbf{Z}^\mathcal{S}$ and $\mathbf{Z}^\mathcal{T}$ be taken as i.i.d and the generalizability of ${\mathcal{M}}_{z}$ be guaranteed.
However, the continuing development of feature-based transfer studies, being popular for a decade and attracting increasing research interests nowadays, presents a contradiction.
On the one hand, existing studies usually claim that they succeed in learning the \emph{domain-invariant} feature representation.
On the other hand, they are quickly defeated by new works on the transfer performance.
Such a `defeat-to-improve' research pattern promotes feature-based transfer studies, but also poses a fact, that is, existing works may fail to achieve the truly \emph{domain-invariant} feature representation.
They do align the two domains, but only yield some suboptimal \emph{domain-invariant} alternatives.

With this in mind, we turn our attention back to ${\mathcal{M}}_{z}$, the transfer model trained using $\mathbf{Z}^\mathcal{S}$.
It is problematic since $\mathbf{Z}^\mathcal{S}$ is unlikely to be truly \emph{domain-invariant}.
Discrepancy still exists between $\mathbf{Z}^\mathcal{S}$ and $\mathbf{Z}^\mathcal{T}$, although it is much smaller than the discrepancy between $\mathbf{X}^\mathcal{S}$ and $\mathbf{X}^\mathcal{T}$.
This degenerates the generalizability of ${\mathcal{M}}_{z}$ to $\mathbf{Z}^\mathcal{T}$, and leaves room to be improved.
Based on this observation, we aim to propose a new transfer model further boosting the performance of ${\mathcal{M}}_{z}$.


\section{The General Form of RTM} \label{RTM_general}
In this section, we present the general form of RTM and discuss how RTM relates to the other popular research topics.

\subsection{Technical Details of RTM}
We start with the training of the conventional transfer model $\mathcal{M}_z$.
By denoting $f_z$ as the prediction function, the learning objective is to minimize the squared error loss as follows:
\begin{equation} \label{eq1}
\mathop{\min}\limits_{f_z} \ \  \mathbb{E}_{\mathbf{z}^\mathcal{S} \sim \mathbf{Z}^\mathcal{S}} [||f_z(\mathbf{z}^\mathcal{S})-\mathbf{y}^\mathcal{S}||_{l_2}] + \alpha \mathcal{R}(f_z),
\end{equation}
where $\mathbb{E}$ is the expectation operator, $\mathbf{y}^\mathcal{S}$ is the corresponding label vector of $\mathbf{z}^\mathcal{S}$, and $\mathcal{R}(\cdot)$ is the regularization term to control the complexity of $f_z$.
As empirically evidenced by existing feature-based transfer methods, $f_z$ can achieve good prediction performance on $\mathbf{Z}^\mathcal{T}$.
Only small region of the new target data are wrongly classified.
If we can take this small region into account in the training, we are expected to achieve improvements on the transfer performance.

To do so, we propose to enlarge the training data population to make it most likely cover the new target data.
More specifically, we propose to randomly corrupt the new source data with some noises, while assigning the artificially corrupted data the same label as the data point from which it is corrupted.
To try the best to cover the new target data, we conduct as many corruptions as possible.
Then, our objective is to train a model $\widetilde{\mathcal{M}}_z$ that performs well on all the possible randomly corrupted data.
This is to minimize the following squared error loss:
\begin{equation} \label{eq2}
\mathop{\min}\limits_{\widetilde{f}_z} \  \mathbb{E}_{\mathbf{z}^\mathcal{S} \sim \mathbf{Z}^\mathcal{S}} [ \ \mathbb{E}_{\boldsymbol\epsilon}[||\widetilde{f}_z(\widetilde{\mathbf{z}}^\mathcal{S})-\mathbf{y}^\mathcal{S}||_{l_2}]\ ] + \alpha \mathcal{R}(\widetilde{f}_z),
\end{equation}
where $\boldsymbol\epsilon$ (under the second expectation operator) is the noise distribution, $\widetilde{\mathbf{z}}^\mathcal{S}$ is the source corrupted data by the noise, and $\widetilde{f}_z$ is the prediction function to be learned.
Compared Eq. (\ref{eq2}) with Eq. (\ref{eq1}), we can see that $\widetilde{\mathcal{M}}_z$ generalizes $\mathcal{M}_z$ by considering much more diverse source data populations.
Assuming we have $J$ different versions of corruptions, then we can rewrite Eq. (\ref{eq2}) as:
\begin{equation} \label{eq3}
\mathop{\min}\limits_{\widetilde{f}_z} \  \frac{1}{nJ} \sum_{i=1}^n \sum_{j=1}^J ||\widetilde{f}_z(\widetilde{\mathbf{z}}_{i,j}^\mathcal{S})-\mathbf{y}_i^\mathcal{S}||_{l_2} + \alpha \mathcal{R}(\widetilde{f}_z),
\end{equation}
where $\widetilde{\mathbf{z}}_{i,j}^\mathcal{S}$ is the $j$-\emph{th} version corruption of $\mathbf{z}_i^\mathcal{S}$.
To simplify, we denote the $J$-times repeated version of $\mathbf{Z}^\mathcal{S}$ as $\widehat{\mathbf{Z}}^\mathcal{S} = [\mathbf{Z}^\mathcal{S},...,\mathbf{Z}^\mathcal{S}]$, the corrupted version of $\widehat{\mathbf{Z}}^\mathcal{S}$ as $\widetilde{\mathbf{Z}}^\mathcal{S}$, and the corresponding $J$-times repeated label matrix as $\widehat{\mathbf{Y}}^\mathcal{S} = [\mathbf{Y}^\mathcal{S},...,\mathbf{Y}^\mathcal{S}]$.
With these notations, Eq. (\ref{eq3}) can be re-formulated as:
\begin{equation} \nonumber
\mathop{\min}\limits_{\widetilde{f}_z} \   \frac{1}{nJ} tr([\widetilde{f}_z(\widetilde{\mathbf{Z}}^\mathcal{S})-\widehat{\mathbf{Y}}^\mathcal{S}] [\widetilde{f}_z(\widetilde{\mathbf{Z}}^\mathcal{S})-\widehat{\mathbf{Y}}^\mathcal{S}]^\mathsf{T}) + \alpha \mathcal{R}(\widetilde{f}_z),
\end{equation}
where $tr$ and $\mathsf{T}$ are trace operator and transpose operator.
We further expand the trace and obtain:
\begin{equation} \nonumber
\mathop{\min}\limits_{\widetilde{f}_z} \   \frac{1}{nJ} tr[\widetilde{f}_z(\widetilde{\mathbf{Z}}^\mathcal{S})\widetilde{f}_z(\widetilde{\mathbf{Z}}^\mathcal{S})^\mathsf{T} - 2 \widetilde{f}_z(\widetilde{\mathbf{Z}}^\mathcal{S})^\mathsf{T} \widehat{\mathbf{Y}}^\mathcal{S} ] + \alpha \mathcal{R}(\widetilde{f}_z).
\end{equation}
We omit $(\widehat{\mathbf{Y}}^\mathcal{S})^\mathsf{T}\widehat{\mathbf{Y}}^\mathcal{S}$ as it does not associate with ${\widetilde{f}_z}$.

The larger $J$ is, the higher the probability of the new target data is covered by the randomly corrupted source data.
The ideal case is conducting infinite corruptions.
However, we cannot make it in practice.
To simulate the infinite random corruptions without really making any corruption,
we propose a marginalization trick.
Specifically, we set $J \to \infty$.
By the weak law of large numbers, this is equivalent to solving the following empirical objective:
\begin{equation} \label{eq4}
\mathop{\min}\limits_{\widetilde{f}_z} \  \frac{1}{n} tr(\mathbb{E}_\epsilon[\widetilde{f}_z(\widetilde{\mathbf{Z}}^\mathcal{S})\widetilde{f}_z(\widetilde{\mathbf{Z}}^\mathcal{S})^\mathsf{T}] - 2 \mathbb{E}_\epsilon[\widetilde{f}_z(\widetilde{\mathbf{Z}}^\mathcal{S})^\mathsf{T}\widehat{\mathbf{Y}}^\mathcal{S}]  ) + \alpha \mathcal{R}(\widetilde{f}_z).
\end{equation}
By specifying the noise and the predictive function, we can solve Eq. (\ref{eq4}) and obtain $\widetilde{f}_z^{opt}$.
We define the model trained following the above procedure as Randomized Transferable Machine (RTM).

\subsection{Relationship with Other Research Topics}
The underlying intuition of RTM is to augment more source data populations in the model learning so that target data can be taken as one of these populations.
However, we highlight that it is significantly different from the existing random data augmentation methods \cite{shorten2019survey}.
Clearly shown in Eq. (\ref{eq4}), we actually do not conduct any random corruption or augmentation, but utilize a marginalization trick to model the ideal case with infinite corruptions.
In the experiment, we have compared with the case where a broad spectrum of actual random corruptions is conducted.
The results show the superiority of the proposed marginalization trick.

RTM also shares the similar idea with the research lines including adversarial training \cite{Dimitris2019Robust} and domain randomization \cite{zakharov2019deceptionnet}.
In adversarial training, a model is trained to be robust to the adversarial examples.
The research efforts mainly confine to finding the adversarial perturbations and to training models that are robust to them.
Domain randomization mainly deals with `sim-to-real' scenarios in reinforcement learning.
It adds noise to simulation environment variables to create more diverse environments, and learns a good policy on all these environments.
The real-world environment is assumed to be included in the constructed environments, and thus the learned policy can be well applied.
Both research lines receive significant research attention, and result in a number of approaches.
This also implies the research potential of RTM.

\section{Instantiations of RTM} \label{Instantiated_RTM}
The above RTM is general as it can be instantiated by using different types of noise distribution and predictive function.
In this section, we propose instantiations of RTM.

\subsection{RTM$_{lr}^d$ with Linear Regression Model}
We first consider the linear regression model as $f_z$.
Specifically, it is formulated as follows:
\begin{equation} \label{eq6}
{f}_z (\mathbf{z}) = \boldsymbol{\omega}^\mathsf{T} \mathbf{z},
\end{equation}
where $\boldsymbol{\omega}$ is the model weights.
By plugging Eq. (\ref{eq6}) into Eq. (\ref{eq4}) and using the $l_2$ norm regularization, we can obtain:
\begin{equation} \label{eq7}
\mathop{\min}\limits_{\widetilde{f}_z} \frac{1}{n} tr(\boldsymbol{\omega}^\mathsf{T}\mathbb{E}_\epsilon[\widetilde{\mathbf{Z}}^\mathcal{S} (\widetilde{\mathbf{Z}}^\mathcal{S})^\mathsf{T}]\boldsymbol{\omega} - 2 \mathbb{E}_\epsilon[ \widehat{\mathbf{Y}}^\mathcal{S}(\widetilde{\mathbf{Z}}^\mathcal{S})^\mathsf{T} ]\boldsymbol{\omega}) + \alpha||\boldsymbol{\omega}||_{2}.
\end{equation}
Note that only $\widetilde{\mathbf{Z}}^\mathcal{S}$ is associated with the noise distribution.
Eq. (\ref{eq7}) is the standard ordinary least squares \cite{Bishop2006}, and it has a closed-form solution as follows:
\begin{equation} \label{eq8}
\boldsymbol{\omega} =( \mathbb{E}_\epsilon[ \widehat{\mathbf{Y}}^\mathcal{S} (\widetilde{\mathbf{Z}}^\mathcal{S})^\mathsf{T} ]) (\mathbb{E}_\epsilon[\widetilde{\mathbf{Z}}^\mathcal{S} (\widetilde{\mathbf{Z}}^\mathcal{S})^\mathsf{T}]+\alpha \mathbf{I})^{-1}.
\end{equation}
where $\mathbf{I}$ is an identity matrix.
Then, our focus in on the two terms $ \mathbb{E}_\epsilon[ \widehat{\mathbf{Y}}^\mathcal{S} (\widetilde{\mathbf{Z}}^\mathcal{S})^\mathsf{T} ]$ and $\mathbb{E}_\epsilon[\widetilde{\mathbf{Z}}^\mathcal{S} (\widetilde{\mathbf{Z}}^\mathcal{S})^\mathsf{T}]$ that are dependent on the noise distribution.
Herein, we consider the widely used dropout noise \cite{wei2018feature}, which is defined as follows:
\begin{Def} \label{def1}
\textbf{\emph{Dropout} noise}: given a data point $\mathbf{x}$, each feature dimension of $\mathbf{x}$ is randomly corrupted by a noise $\epsilon$ that draws a Bernoulli distribution with a probability $p$. That is to say, each feature is corrupted to 0 with the probability $p$ and retains with the probability $1-p$.
\end{Def}

With the dropout noise, we can derive the two terms.
Specifically, for $ \mathbb{E}_\epsilon[ \widehat{\mathbf{Y}}^\mathcal{S} (\widetilde{\mathbf{Z}}^\mathcal{S})^\mathsf{T} ]$ we have:
\begin{equation} \label{eq9}
 \mathbb{E}_\epsilon[ \widehat{\mathbf{Y}}^\mathcal{S} (\widetilde{\mathbf{Z}}^\mathcal{S})^\mathsf{T} ]= (1-p) \mathbf{Y}^\mathcal{S}(\mathbf{Z}^\mathcal{S})^\mathsf{T}.
\end{equation}
For $\mathbb{E}_\epsilon[\widetilde{\mathbf{Z}}^\mathcal{S} (\widetilde{\mathbf{Z}}^\mathcal{S})^\mathsf{T}]$, when $\alpha \ne \beta$,
\begin{equation} \label{eq10}
[\mathbb{E}_\epsilon[\widetilde{\mathbf{Z}}^\mathcal{S} (\widetilde{\mathbf{Z}}^\mathcal{S})^\mathsf{T}]]_{\alpha,\beta} =
 (1-p)^2[\mathbf{Z}^\mathcal{S} (\mathbf{Z}^\mathcal{S})^\mathsf{T}]_{\alpha,\beta},
\end{equation}
and when $\alpha = \beta$,
\begin{equation} \label{eq11}
\begin{split}
[\mathbb{E}_\epsilon[\widetilde{\mathbf{Z}}^\mathcal{S} (\widetilde{\mathbf{Z}}^\mathcal{S})^\mathsf{T}]]_{\alpha,\alpha} =
 (1-p)[\mathbf{Z}^\mathcal{S} (\mathbf{Z}^\mathcal{S})^\mathsf{T}]_{\alpha,\alpha},
\end{split}
\end{equation}
where subscripts $\alpha$ and $\beta$ are the row index and the column index of a matrix, respectively.
With the above equations, we can obtain the model weights by solving Eq. (\ref{eq8}).
We define such an instantiation as RTM$_{lr}^d$ where the subscript indicates the linear regression model and the superscript indicates the dropout noise.
The learning procedure of RTM$_{lr}^d$ is summarized in Algorithm \ref{rtm1}.
Notably, RTM$_{lr}^d$ can be implemented by several lines of code, and very efficient to be trained.
\begin{algorithm}[t]
  \caption{ RTM$_{lr}^d$ with Linear Logistic Regression Model and Dropout Noise }
  \label{rtm1}
  \begin{algorithmic}
    \STATE {\textbf{Input:}  Source data matrix with new feature representation ${\mathbf{Z}}^\mathcal{S}$, source label matrix ${\mathbf{Y}}_\mathcal{S}$, and the corruption probability in dropout noise $p$.}
    \STATE {\ \ \ 1. Calculate  $ \mathbb{E}_\epsilon[ \widehat{\mathbf{Y}}^\mathcal{S} (\widetilde{\mathbf{Z}}^\mathcal{S})^\mathsf{T} ]$ using Eq. (\ref{eq9});}
    \STATE {\ \ \ 2. Calculate $\mathbb{E}_\epsilon[\widetilde{\mathbf{Z}}^\mathcal{S} (\widetilde{\mathbf{Z}}^\mathcal{S})^\mathsf{T}]$ using Eqs. (\ref{eq10}) and (\ref{eq11});}
    \STATE {\ \ \ 3. Calculate the model weight matrix $\boldsymbol{\omega}$ using Eq. (\ref{eq8})}.
    \STATE \textbf{Output:} {The prediction function $\widetilde{f}_z$, i.e., the transfer model $\mathcal{M}_z$.}
  \end{algorithmic}
\end{algorithm}

\subsection{RTM$_{kr}^d$ with Kernel Ridge Regression Model}
RTM$_{lr}^d$ only captures the data linearity.
To explore the nonlinearity, we can use the kernel ridge regression model as $f_z$.
Specifically, it is formulated as:
\begin{equation} \nonumber
{f}_{k_z} (\mathbf{z}) = \boldsymbol{\omega}_k^\mathsf{T} \mathbf{k}_{\mathbf{z}},
\end{equation}
where $\mathbf{k}_{\mathbf{z}}$ is a kernel vector calculated by using a predefined kernel function $k(\cdot,\cdot)$ on $\mathbf{z}$.
Similarly, we have:
\begin{equation} \nonumber
\begin{aligned}
\mathop{\min}\limits_{\widetilde{f}_{k_z}} &\frac{1}{n} tr(\boldsymbol{\omega}_k^\mathsf{T}\mathbb{E}_\epsilon[\widetilde{\mathbf{K}}_{\mathbf{z}}^\mathcal{S} (\widetilde{\mathbf{K}}_{\mathbf{z}}^\mathcal{S})^\mathsf{T}]\boldsymbol{\omega}_k - 2 \mathbb{E}_\epsilon[ \widehat{\mathbf{Y}}^\mathcal{S}(\widetilde{\mathbf{K}}_{\mathbf{z}}^\mathcal{S})^\mathsf{T} ]\boldsymbol{\omega}_k ) \\
&+ \alpha||\boldsymbol{\omega}_k||_{2}.
\end{aligned}
\end{equation}
where $\widetilde{\mathbf{K}}_{\mathbf{z}}^\mathcal{S}$ is the corrupted version of the duplicated ${\mathbf{K}}_{\mathbf{z}}^\mathcal{S}$.
The closed-form solution is:
\begin{equation} \nonumber
\boldsymbol{\omega}_k =( \mathbb{E}_\epsilon[ \widehat{\mathbf{Y}}^\mathcal{S} (\widetilde{\mathbf{K}}_{\mathbf{z}}^\mathcal{S})^\mathsf{T} ]) (\mathbb{E}_\epsilon[\widetilde{\mathbf{K}}_{\mathbf{z}}^\mathcal{S} (\widetilde{\mathbf{K}}_{\mathbf{z}}^\mathcal{S})^\mathsf{T}]+\alpha \mathbf{I})^{-1}.
\end{equation}
By using the above dropout noise, we can derive:
\begin{equation} \nonumber
 \mathbb{E}_\epsilon[ \widehat{\mathbf{Y}}^\mathcal{S} (\widetilde{\mathbf{K}}_{\mathbf{z}}^\mathcal{S})^\mathsf{T} ]= (1-p) \mathbf{Y}^\mathcal{S}(\mathbf{K}_{\mathbf{z}}^\mathcal{S})^\mathsf{T}.
\end{equation}
For $\mathbb{E}_\epsilon[\widetilde{\mathbf{K}}_{\mathbf{z}}^\mathcal{S} (\widetilde{\mathbf{K}}_{\mathbf{z}}^\mathcal{S})^\mathsf{T}]$, when $\alpha \ne \beta$,
\begin{equation} \nonumber
[\mathbb{E}_\epsilon[\widetilde{\mathbf{K}}_{\mathbf{z}}^\mathcal{S} (\widetilde{\mathbf{K}}_{\mathbf{z}}^\mathcal{S})^\mathsf{T}]]_{\alpha,\beta} =
 (1-p)^2[\mathbf{K}_{\mathbf{z}}^\mathcal{S} (\mathbf{K}_{\mathbf{z}}^\mathcal{S})^\mathsf{T}]_{\alpha,\beta},
\end{equation}
and when $\alpha = \beta$,
\begin{equation} \nonumber
\begin{split}
[\mathbb{E}_\epsilon[\widetilde{\mathbf{K}}_{\mathbf{z}}^\mathcal{S} (\widetilde{\mathbf{K}}_{\mathbf{z}}^\mathcal{S})^\mathsf{T}]]_{\alpha,\alpha} =
 (1-p)[\mathbf{K}_{\mathbf{z}}^\mathcal{S} (\mathbf{K}_{\mathbf{z}}^\mathcal{S})^\mathsf{T}]_{\alpha,\alpha},
\end{split}
\end{equation}

As can be seen, mathematically, RTM$_{kr}^d$ is similar as RTM$_{lr}^d$. RTM$_{kr}^d$ degenerates to RTM$_{lr}^d$ if a linear kernel is applied.
In the following sections, we focus on RTM$_{lr}^d$ to demonstrate the effectiveness of RTM.

\subsection{Transfer Effectiveness Analysis of RTM$_{lr}^d$}
In this section, we theoretically show how RTM$_{lr}^d$ differs from the the conventional transfer model, and interpret the underlying rationale on the transfer effectiveness of RTM$_{lr}^d$ from the perspective of adaptive regularization \cite{wager2013dropout}.

We reformulate the learning objective of RTM$_{lr}^d$ by plugging Eqs. (\ref{eq9}-\ref{eq11}) into Eq. (\ref{eq7}), and the we obtain:
\begin{equation} \label{eq12}
\begin{aligned}
\mathop{\min}\limits_{\widetilde{f}_z} &\frac{1}{n} tr(\boldsymbol{\omega}^\mathsf{T}[(1-p)^2\widehat{\mathbf{Z}}^\mathcal{S}(\widehat{\mathbf{Z}}^\mathcal{S})^\mathsf{T} - (1-p)^2\mathbf{\Omega}+(1-p)\mathbf{\Omega}]\boldsymbol{\omega}) \\&- 2(1-p)\widehat{\mathbf{Y}}^\mathcal{S}(\widehat{\mathbf{Z}}^\mathcal{S})^\mathsf{T} \boldsymbol{\omega}) + \alpha||\boldsymbol{\omega}||_{2},
\end{aligned}
\end{equation}
where $\mathbf{\Omega}$ is a diagonal matrix whose diagonal entry is the same as $\widehat{\mathbf{Z}}^\mathcal{S}(\widehat{\mathbf{Z}}^\mathcal{S})^\mathsf{T}$.
We reorganize Eq. (\ref{eq12}):
\begin{equation} \label{eq13}
\begin{aligned}
\mathop{\min}\limits_{\widetilde{f}_z} & \underbrace{\frac{1}{n} tr(\boldsymbol{\omega}^\mathsf{T}\widehat{\mathbf{Z}}^\mathcal{S}(\widehat{\mathbf{Z}}^\mathcal{S})^\mathsf{T}\boldsymbol{\omega} - 2 \widehat{\mathbf{Y}}^\mathcal{S}(\widehat{\mathbf{Z}}^\mathcal{S})^\mathsf{T} \boldsymbol{\omega}) + \alpha||\boldsymbol{\omega}||_{2}}_{Learning \ objective \ of \ {\mathcal{M}}_{z}}
 \\ &\underbrace{+\frac{1}{n} tr(\boldsymbol{\omega}^\mathsf{T}[(p^2-2p)\widehat{\mathbf{Z}}^\mathcal{S}(\widehat{\mathbf{Z}}^\mathcal{S})^\mathsf{T} -(p^2-p)\mathbf{\Omega}]\boldsymbol{\omega})}_{Task-dependent \ Frobenius \ constraint}
\\&\underbrace{+ \frac{2p}{n} \times tr(\widehat{\mathbf{Y}}^\mathcal{S}(\widehat{\mathbf{Z}}^\mathcal{S})^\mathsf{T} \boldsymbol{\omega}).}_{Task-dependent \  low-rank \ constraint}
\end{aligned}
\end{equation}
From Eq. (\ref{eq13}), we can see that the learning objective of RTM$_{lr}^d$ consists of three parts: the learning objective of the conventional transfer model plus two task-dependent\footnote{Task-dependent refers to the data and the noise used in the transfer task.} regularization terms.
Eq. (\ref{eq13}) clearly shows the difference of RTM$_{lr}^d$ and the conventional transfer model, i.e., the data-dependent regularization terms.
We thus induce that it is the two terms\footnote{Existing transfer methods, e.g., \cite{li2017domain}, have shown the transfer effectiveness of the low-rank constrained model. Although our case is not the standard low-rank norm, we believe the data-dependent constraints also benefits the transfer as also empirically verified by the empirical results. } that lead to the transfer superiority of RTM over the conventional transfer model.

\subsection{Discussions on RTM$_{lr}^d$} \label{S4}
\subsubsection{More Advanced Instantiations}
For RTM$_{lr}^d$, it is a simple instantiation of RTM.
More advanced instantiations can be expected by using other types of noise, e.g., multinomial dropout noise, and predictive function, e.g., support vector machine.
In particular, if we still exploit linear regression model but use another type of noise, then the problem is to adapt $ \mathbb{E}_\epsilon[ \widehat{\mathbf{Y}}^\mathcal{S} (\widetilde{\mathbf{Z}}^\mathcal{S})^\mathsf{T} ]$ and $\mathbb{E}_\epsilon[\widetilde{\mathbf{Z}}^\mathcal{S} (\widetilde{\mathbf{Z}}^\mathcal{S})^\mathsf{T}]$ to this new type of noise.
With different noise distributions and predictive functions, Eq. (\ref{eq4}) has different forms, and thus the optimization would be different.
\emph{We re-emphasize that, in this work, our focus is to present the idea of RTM, and show its transfer effectiveness through a simple instantiation.}
We start this new research topic for transfer learning, and leave extended instantiations in future studies.

\subsubsection{Computational Complexity}
One advantage of RTM$_{lr}^d$ is its fast and efficient training process.
As shown in Eqs. (\ref{eq7}) and (\ref{eq8}), it is to solve a $k \times k$ system of linear equations, where $k$ is the feature dimensionality of the new feature representation.
This can be solved by a non-iterative closed-form solution where the computational complexity is $\mathcal{O}(k^3)$.
When the new data is obtained from subspace-based transfer methods, $k$ is usually much smaller than the original feature dimensionality, which enables a considerably low computational cost.
Such computation bonus is discounted if the new data is obtained from deep learning based transfer methods where much more new features are generated.
Thus, when evaluating RTM$_{lr}^d$, we tend to use subspace-based methods to generate new data.


\subsubsection{Technical Significance}
Technically, the learning objective of RTM$_{lr}^d$, i.e., Eq. (\ref{eq7}), has similar form with that of \cite{chen2012marginalized,clinchant2016domain,Wei2016DeepNF,wei2018feature}.
However, a major difference between our work with these works is that these works are feature-based transfer methods that are dedicated to learning the new feature representation, while our work focuses on a new transfer model.
In another words, those methods aim to learn $\mathbf{Z}^\mathcal{S}$ and $\mathbf{Z}^\mathcal{T}$, and our work aims to learn $\widetilde{\mathcal{M}}_z$.
Note that \cite{chen2012marginalized,clinchant2016domain,Wei2016DeepNF,wei2018feature} pose a problem on how to set the corruption probability $p$, and exploit a cross-validation strategy to handle it.
Our RTM$_{lr}^d$ also faces such a problem.
However, it is not our focus and we can handle it using the existing strategy.

\subsubsection{Scalability on Feature-based Transfer Methods}
RTM$_{lr}^d$ is built on $\mathbf{Z}^\mathcal{S}$, which can be learned by either subspace learning or deep feature learning.
Situation differs slightly between these two feature learning strategies.
For subspace learning methods, RTM$_{lr}^d$ can be seamlessly followed up as these methods purely focus on the new feature representation learning.
For deep feature learning methods that couple the model training with the feature learning, we have two options.
One option is to extract out the new data after the whole deep learning process, and then apply RTM$_{lr}^d$.
Another option is to link up RTM$_{lr}^d$ with the feature learning, i.e., take Eq. (\ref{eq4}) as a part of the objective of existing deep feature learning methods.

\subsubsection{Applicability to the Original Data}
Although RTM is proposed to build upon the new data, it can be readily adapted to the original data.
We denote RTM trained on $\mathbf{X}^\mathcal{S}$ as $\widetilde{\mathcal{M}}_x$.
We are curious about the transfer capability of $\widetilde{\mathcal{M}}_x$, and thus we conduct the empirical comparisons of $\widetilde{\mathcal{M}}_x$ with $\mathcal{M}_x$, the transfer model trained on $\mathbf{X}^\mathcal{S}$.
As expected, we observe that $\widetilde{\mathcal{M}}_x$ consistently achieves better transfer performance than $\mathcal{M}_x$.
This again verifies the effectiveness of the idea of RTM.

Moreover, $\widetilde{\mathcal{M}}_x$ can be regarded as a new independent transfer method as it directly works on the original data.
We thus compare $\widetilde{\mathcal{M}}_x$ with state-of-the-art feature-based transfer methods including both subspace-based and deep learning based ones, i.e., $\mathcal{M}_z$ with the new data learned by these methods.
Surprisingly, we find that $\widetilde{\mathcal{M}}_x$ even outperforms $\mathcal{M}_z$ in some transfer tasks, which further indicates RTM is a promising transfer model.
\begin{figure}[t]
\centering
{\includegraphics[scale = 0.45]{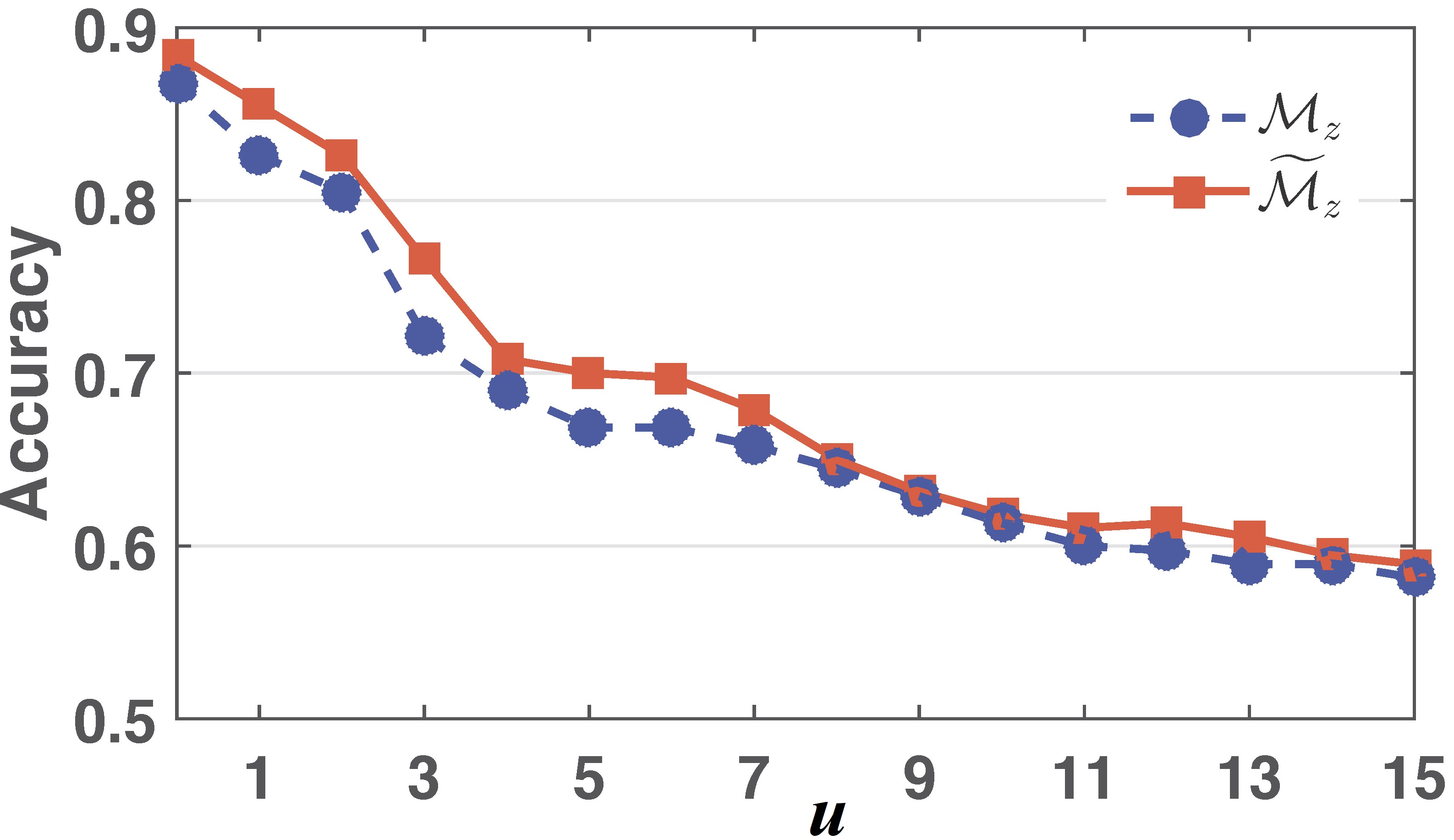} }
 \caption{Results on synthetic datasets.
 } \label{artificial}
\end{figure}

\section{Empirical Evaluations}
In this section, we conduct empirical studies on both synthetic and real-world datasets.
We mainly verify the transfer superiority of $\widetilde{\mathcal{M}}_z$ to $\mathcal{M}_z$.
We also test the performance of RTM$_{lr}^d$ on the original data by comparing $\widetilde{\mathcal{M}}_x$ with $\mathcal{M}_x$.
Property studies including model similarity analysis and data augmentation analysis are also done.
All the experiments are done on a 64-bit operating system with Intel(R) Xeon(R) CPU E5-1650 0 @ 3.20GHz using Matlab 2019b.

\subsection{Synthetic Experiments} \label{S5-1}
We first evaluate the proposed RTM$_{lr}^d$ on a synthetic dataset to show the transfer superiority of $\widetilde{\mathcal{M}}_z$ to $\mathcal{M}_z$.
A broad spectrum of transfer tasks with the domain divergence of the source and target domains varying from low to high is artificially constructed.
Precisely, we generate target data by a linear function $f(\mathbf{z}) = \mathbf{w}_0^{\rm T}{\mathbf{z}}+\sigma $, where $\mathbf{w}_0 \in \mathbb{R}^{100}$ and $\sigma$ is a zero-mean Gaussian noise term.
We randomly sample 380 input data points drawn a Gaussian distribution and then use $f$ to generate their output values.
A sigmoid function $sig(y) = e^{y}/(1+e^{y})-0.5$ and a sign function are then used to transfer the output value to be binary.
For the source domain, we randomly sample another 380 input data points, and then use $g(\mathbf{z}) = (\mathbf{w}_0^{\rm T}+\mu\Delta\mathbf{w}){\mathbf{z}}+\sigma $, where $\Delta\mathbf{w}$ is a random fluctuation vector and $\mu$ is the variable controlling the divergence between $f$ and $g$, to generate 380 points for each source with different $\mu$.
Note that $\mu$ takes value from 0 to 15 with the step as 1, and a higher $\mu$ indicates a higher domain divergence.
By doing so, we construct 16 transfer tasks.
\begin{table*}[!t]
\caption{Comparisons of $\widetilde{\mathcal{M}}_z$ and $\mathcal{M}_z$ on the feature representation of subspace-based transfer methods}
    \label{compaful}
\resizebox{\linewidth}{!}{
\renewcommand\arraystretch{1}
\begin{tabular}{|c||c|c||c|c||c|c||c|c||c|c||c|c||c|c||c|c|}
\hline
Task& \multicolumn{2}{|c||}{TCA} & \multicolumn{2}{c||}{GFK} & \multicolumn{2}{c||}{SA} & \multicolumn{2}{c||}{TJM} & \multicolumn{2}{c||}{JDA} & \multicolumn{2}{c||}{CORAL} & \multicolumn{2}{c||}{JGSA} & \multicolumn{2}{c|}{MMIT} \\ \hline
Model   & $\mathcal{M}_z$        & $\widetilde{\mathcal{M}}_z$        & $\mathcal{M}_z$        & $\widetilde{\mathcal{M}}_z$         & $\mathcal{M}_z$       & $\widetilde{\mathcal{M}}_z$         & $\mathcal{M}_z$        & $\widetilde{\mathcal{M}}_z$         & $\mathcal{M}_z$        & $\widetilde{\mathcal{M}}_z$         & $\mathcal{M}_z$         & $\widetilde{\mathcal{M}}_z$          & $\mathcal{M}_z$        & $\widetilde{\mathcal{M}}_z$          & $\mathcal{M}_z$        & $\widetilde{\mathcal{M}}_z$          \\ \hline \hline
C-A     & 53.44       & \textcolor{red}{53.65}      & \textcolor{red}{55.32}       & 54.07      & 28.29      & \textcolor{red}{46.24}      & 51.77       & \textcolor{red}{51.77}      & 49.27       & \textcolor{red}{49.69}      & 35.07        & \textcolor{red}{54.18}       & 51.67       & \textcolor{red}{54.28}       & 51.98       & \textcolor{red}{52.82}       \\ \hline
C-W     & 50.51       & \textcolor{red}{51.86}      & 49.49       & \textcolor{red}{51.19}      & 23.39      & \textcolor{red}{37.63}      & 49.49       &  \textcolor{red}{49.49 }     & 48.14       &  \textcolor{red}{50.17}      & 31.19        &  \textcolor{red}{48.14}       & 45.42       &  \textcolor{red}{49.49}       & 49.15       &  \textcolor{red}{50.85}       \\ \hline
C-D     & 47.77       &  \textcolor{red}{49.04 }     & 43.95       &  \textcolor{red}{49.04}      & 21.66      &  \textcolor{red}{45.22}      & 41.40       &  \textcolor{red}{42.04}      & 47.13       &  \textcolor{red}{49.68}      & 31.85        &  \textcolor{red}{46.50}       & 40.76       & \textcolor{red}{ 45.86}       & 47.13       &  \textcolor{red}{48.41}       \\ \hline
A-C     & 42.83       &  \textcolor{red}{43.54}      & 41.94       &  \textcolor{red}{44.52}      & 26.54      &  \textcolor{red}{39.63}      &  \textcolor{red}{44.97}       & 44.70      & 38.91       &  \textcolor{red}{42.39}      & 28.50        &  \textcolor{red}{43.72}       & 42.65       &  \textcolor{red}{44.17}       & 42.56       &  \textcolor{red}{43.81}       \\ \hline
A-W     & 38.98       &  \textcolor{red}{41.02}      & 40.68       &  \textcolor{red}{43.73}      & 16.27      &  \textcolor{red}{36.61}      & 45.08       &  \textcolor{red}{45.76}      & 46.78       &  \textcolor{red}{48.81}      & 25.42        &  \textcolor{red}{41.69}       & 35.59       &  \textcolor{red}{41.36}       & 40.00       &  \textcolor{red}{41.02}       \\ \hline
A-D     & 40.13       &  \textcolor{red}{43.31}      & 38.85       &  \textcolor{red}{48.41}      & 23.57      &  \textcolor{red}{39.49}      & 43.95       &  \textcolor{red}{44.59}      & 48.41       &  \textcolor{red}{48.41}      & 28.66        &  \textcolor{red}{43.31}       & 37.58       &  \textcolor{red}{42.68}       & 41.40       &  \textcolor{red}{43.31}       \\ \hline
W-C     & 36.69       &  \textcolor{red}{37.22}      & 36.69       &  \textcolor{red}{36.87}      & 24.49      &  \textcolor{red}{34.28}      &  \textcolor{red}{39.36}       & 39.09      & 32.32       &  \textcolor{red}{33.30}      & 30.01        &  \textcolor{red}{35.80}       & 34.55       &  \textcolor{red}{35.17}       &  \textcolor{red}{38.11}       & 37.13       \\ \hline
W-A     & 40.29       &  \textcolor{red}{40.50}      & 37.79       &  \textcolor{red}{40.81}      & 19.73      &  \textcolor{red}{37.47}      & 39.98       &  \textcolor{red}{40.19}      &  \textcolor{red}{36.12}       & 35.91      & 33.92        &  \textcolor{red}{39.87}       & 38.31       &  \textcolor{red}{39.46}       & 40.29       &  \textcolor{red}{40.40}       \\ \hline
W-D     & 77.71       &  \textcolor{red}{78.98}      & 74.52       &  \textcolor{red}{78.34}      & 41.40      &  \textcolor{red}{65.61}      & 70.06       &  \textcolor{red}{70.70}      & 74.52       &  \textcolor{red}{75.80}      & 77.71        &  \textcolor{red}{84.08}       & 82.80       &  \textcolor{red}{83.44}       & 76.43       &  \textcolor{red}{78.98}       \\ \hline
D-C     & 35.62       &  \textcolor{red}{36.60}      & 30.45       &  \textcolor{red}{36.78}      & 20.66      &  \textcolor{red}{35.89}      & 33.13       &  \textcolor{red}{33.93}      &  \textcolor{red}{32.50}       & 32.32      & 29.74        &  \textcolor{red}{35.53}       & 32.95       &  \textcolor{red}{33.48}       &  \textcolor{red}{35.71}       & 34.73       \\ \hline
D-A     & 38.83       &  \textcolor{red}{38.83}      & 40.40       &  \textcolor{red}{41.02}      & 26.62      &  \textcolor{red}{36.01}      &  \textcolor{red}{35.18}       & 35.07      &  \textcolor{red}{35.59}       & 35.39      & 33.72        &  \textcolor{red}{38.10}       & 35.91       &  \textcolor{red}{36.12}       &  \textcolor{red}{39.35}       & 38.10       \\ \hline
D-W     & 80.34       &  \textcolor{red}{82.03}      & 73.90       &  \textcolor{red}{80.00}      & 37.29      &  \textcolor{red}{64.75}      & 74.58       &  \textcolor{red}{76.27}      & 77.63       &  \textcolor{red}{77.63}      & 80.00        &  \textcolor{red}{85.08}       & 84.07       &  \textcolor{red}{84.07}       & 79.32       &  \textcolor{red}{79.66}       \\ \hline
Mean & 48.60       &  \textcolor{red}{49.72}      & 47.00       &  \textcolor{red}{50.40}      & 25.82      &  \textcolor{red}{43.24}      & 47.41       &  \textcolor{red}{47.80}      & 47.28       &  \textcolor{red}{48.29}      & 38.82        &  \textcolor{red}{49.67}       & 46.86       &  \textcolor{red}{49.13}       & 48.45       &  \textcolor{red}{49.10}       \\ \hline \hline
CL$_{12}$  & 77.36       &  \textcolor{red}{80.42}      & 62.22       &  \textcolor{red}{78.47}      & 48.06      &  \textcolor{red}{76.11}      & 74.72       &  \textcolor{red}{76.11}      & 83.06       &  \textcolor{red}{83.06}      & 68.47        &  \textcolor{red}{79.58}       & 76.53       &  \textcolor{red}{79.03}       & 76.94       &  \textcolor{red}{81.81}       \\ \hline
CL$_{21}$  & 78.06       &  \textcolor{red}{78.61}      & 62.92       &  \textcolor{red}{77.78}      & 51.67      &  \textcolor{red}{74.58}      & 76.81       &  \textcolor{red}{77.50}      & 83.06       & \textcolor{red}{ 83.19}      & 68.06        &  \textcolor{red}{78.19}       & 73.89       &  \textcolor{red}{76.39}       & 77.08       &  \textcolor{red}{79.44}       \\ \hline
Mean & 77.71       &  \textcolor{red}{79.51}      & 62.57       &  \textcolor{red}{78.13}      & 49.86      &  \textcolor{red}{75.35}      & 75.76       &  \textcolor{red}{76.81}      & 83.06       &  \textcolor{red}{83.13}      & 68.26        &  \textcolor{red}{78.89}       & 75.21       &  \textcolor{red}{77.71}       & 77.01       &  \textcolor{red}{80.63}       \\ \hline \hline
CR-1    & 68.29       &  \textcolor{red}{71.72}      & 61.95       &  \textcolor{red}{73.01}      & 62.04      &  \textcolor{red}{71.98}      &  \textcolor{red}{76.69}       & 74.81      & 64.52       &  \textcolor{red}{66.41}      & 60.50        &  \textcolor{red}{69.58}       & 63.24       &  \textcolor{red}{69.58}       & 71.12       &  \textcolor{red}{71.72}       \\ \hline
CR-2    &  \textcolor{red}{81.42}      & 81.34      & 75.74       &  \textcolor{red}{79.30}      &  \textcolor{red}{76.42}      & 73.45      & 74.13       &  \textcolor{red}{74.55}      & 81.51       &  \textcolor{red}{83.21}      & 74.89        &  \textcolor{red}{84.22}       & 83.04       &  \textcolor{red}{84.22}       & 81.51       &  \textcolor{red}{82.02}       \\ \hline
CS-1    &  \textcolor{red}{73.48}       & 73.40      & 70.15       &  \textcolor{red}{83.92}      & 68.69      &  \textcolor{red}{83.92}      &  \textcolor{red}{78.96}       & 76.99      & 87.17       &  \textcolor{red}{87.34}      & 68.35        &  \textcolor{red}{78.61}       & 77.16       &  \textcolor{red}{78.61}       &  \textcolor{red}{78.44}       & 77.59       \\ \hline
CS-2    &  \textcolor{red}{70.44}       & 69.85      & 73.68       &  \textcolor{red}{79.30}      & 72.91      &  \textcolor{red}{78.96}      &  \textcolor{red}{73.94}       & 73.59      & 73.25       &  \textcolor{red}{73.85}      & 70.19        &  \textcolor{red}{75.72}       & 72.74       &  \textcolor{red}{75.72}       & 71.29       &  \textcolor{red}{77.27}       \\ \hline
CT-1    &  \textcolor{red}{80.48}       & 79.43      & 64.98       &  \textcolor{red}{85.65}      & 63.83      &  \textcolor{red}{80.29}      &  \textcolor{red}{82.87}       & 82.49      &  \textcolor{red}{86.12}       & 85.93      & 63.54        &  \textcolor{red}{90.05}       & 86.89       &  \textcolor{red}{90.05}       &  \textcolor{red}{86.22}       & 85.84       \\ \hline
CT-2    & 84.33       &  \textcolor{red}{84.96}      & 81.39       &  \textcolor{red}{87.17}      & 78.97      &  \textcolor{red}{84.12}      & 84.54       &  \textcolor{red}{86.01}      & 88.12       &  \textcolor{red}{88.43}      & 81.91        &  \textcolor{red}{90.12}       & 89.70       &  \textcolor{red}{90.12}       &  \textcolor{red}{88.12}       & 87.49       \\ \hline
RS-1    &  \textcolor{red}{74.77}       & 74.68      & 71.91       &  \textcolor{red}{77.88}      & 71.07      &  \textcolor{red}{73.17}      & 79.14       &  \textcolor{red}{79.73}      & 78.64       &  \textcolor{red}{79.39}      & 70.06        &  \textcolor{red}{75.36}       & 70.73       &  \textcolor{red}{75.36}       & 72.83       &  \textcolor{red}{73.25}       \\ \hline
RS-2    & 74.79       &  \textcolor{red}{75.04}      & 70.99       &  \textcolor{red}{76.90}      & 72.43      &  \textcolor{red}{76.81}      &  \textcolor{red}{74.37}       & 73.95      & 81.45       &  \textcolor{red}{81.79}      & 72.26        &  \textcolor{red}{79.93}       & 79.09       &  \textcolor{red}{79.93}       &  \textcolor{red}{75.13}       & 74.79       \\ \hline
RT-1    &  \textcolor{red}{68.28}       & 67.99      & 60.42       &  \textcolor{red}{67.42}      & 59.56      &  \textcolor{red}{64.02}      & 76.61       &  \textcolor{red}{76.61}      &  \textcolor{red}{75.57}       & 75.19      & 60.98        &  \textcolor{red}{68.84}       & 64.11       &  \textcolor{red}{68.84}       &  \textcolor{red}{77.37}       & 76.61       \\ \hline
RT-2    & 77.52       &  \textcolor{red}{77.82}      & 67.25       &  \textcolor{red}{73.20}      & 68.07      &  \textcolor{red}{69.82}      &  \textcolor{red}{79.57}       & 79.36      & 81.11       &  \textcolor{red}{81.31}      & 69.82        &  \textcolor{red}{75.98}       & 67.15       &  \textcolor{red}{75.98}       &  \textcolor{red}{80.80}       & 79.98       \\ \hline
ST-1    & 69.82       &  \textcolor{red}{69.91}    & 59.98       &  \textcolor{red}{69.91}      & 59.22      &  \textcolor{red}{62.25}      & 79.19       &  \textcolor{red}{79.28}      &  \textcolor{red}{69.06}       & 68.31      & 58.85        &  \textcolor{red}{79.85}       & 77.29       &  \textcolor{red}{79.85}       &  \textcolor{red}{73.60}       & 73.42       \\ \hline
ST-2    &  \textcolor{red}{82.01}       & 81.80      & 66.18       &  \textcolor{red}{75.90}      & 64.74      &  \textcolor{red}{71.25}      & 82.32       &  \textcolor{red}{83.14}      & 82.94       &  \textcolor{red}{84.28}      & 67.11        &  \textcolor{red}{86.14}       & 83.66       &  \textcolor{red}{86.14}       & 82.21       &  \textcolor{red}{83.28}       \\ \hline
Mean & 75.47       &  \textcolor{red}{75.66}      & 68.72       &  \textcolor{red}{77.46 }     & 68.16      &  \textcolor{red}{74.17}      &  \textcolor{red}{78.53}       & 78.38      & 79.12       &  \textcolor{red}{79.62}      & 68.21        &  \textcolor{red}{79.53}       & 76.23       &  \textcolor{red}{79.53}       & 78.22       &  \textcolor{red}{78.44 }      \\ \hline \hline
B-D     & 77.64       &  \textcolor{red}{78.49}      &  \textcolor{red}{79.64}       & 79.49      & 71.99      &  \textcolor{red}{76.74}      & 76.89       &  \textcolor{red}{77.19}      & 76.39       &  \textcolor{red}{76.54}      & 76.04        &  \textcolor{red}{79.49}       & 78.69       &  \textcolor{red}{78.84}       & 78.49       &  \textcolor{red}{79.34}       \\ \hline
B-E     & 76.48       &  \textcolor{red}{76.48}      & 78.23       &  \textcolor{red}{78.28}      & 67.67      &  \textcolor{red}{75.88}      &  \textcolor{red}{79.73}       & 79.58      & 78.33       &  \textcolor{red}{78.38}      & 72.52        &  \textcolor{red}{79.03}       & 77.03       &  \textcolor{red}{77.28}       &  \textcolor{red}{75.48}       & 75.38       \\ \hline
B-K     &  \textcolor{red}{78.19}       & 78.14      & 79.24       &  \textcolor{red}{80.64}      & 60.68      &  \textcolor{red}{76.64}      & 79.09       &  \textcolor{red}{79.14}      & 78.84       &  \textcolor{red}{80.04}      & 74.04        &  \textcolor{red}{80.84}       & 79.39       &  \textcolor{red}{79.39}       & 76.99       &  \textcolor{red}{77.74}       \\ \hline
D-B     & 78.05       &  \textcolor{red}{78.60}      & 78.05       &  \textcolor{red}{79.20}      & 66.50      &  \textcolor{red}{75.70}      &  \textcolor{red}{78.05}       & 77.60      &  \textcolor{red}{76.60}       & 76.45      & 75.90        &  \textcolor{red}{79.90}       & 78.80       &  \textcolor{red}{79.85}       & 77.80       &  \textcolor{red}{78.30}       \\ \hline
D-E     &  \textcolor{red}{78.18}       & 78.13      & 77.63       &  \textcolor{red}{80.08}      & 70.87      &  \textcolor{red}{77.48}      & 79.53       &  \textcolor{red}{79.73}      & 80.48       &  \textcolor{red}{80.48}      & 69.57        &  \textcolor{red}{78.98}       & 76.68       &  \textcolor{red}{77.38}       &  \textcolor{red}{77.58}       & 76.98       \\ \hline
D-K     &  \textcolor{red}{78.44}       & 78.29      & 78.84       &  \textcolor{red}{80.64}      & 65.93      &  \textcolor{red}{79.19}      &  \textcolor{red}{79.54}       & 79.29      & 79.49       &  \textcolor{red}{79.74}      & 72.54        &  \textcolor{red}{81.34}       & 79.44       &  \textcolor{red}{80.29}       & 78.79       &  \textcolor{red}{78.84}       \\ \hline
E-B     & 73.75       &  \textcolor{red}{73.85}      & 73.70       &  \textcolor{red}{75.05}      & 58.75      &  \textcolor{red}{71.30}      & 72.55       &  \textcolor{red}{73.65}      & 73.60       &  \textcolor{red}{73.60}      & 70.25        &  \textcolor{red}{75.55}       & 74.45       &  \textcolor{red}{74.65}       & 72.75       &  \textcolor{red}{73.35}       \\ \hline
E-D     & 74.84       &  \textcolor{red}{74.99}      & 74.34       &  \textcolor{red}{76.74}      & 62.68      &  \textcolor{red}{74.54}      & 75.49       &  \textcolor{red}{75.69}      &  \textcolor{red}{75.64}       & 75.44      & 69.88        &  \textcolor{red}{74.94}       & 74.09       &  \textcolor{red}{74.34}       & 74.39       &  \textcolor{red}{74.84}       \\ \hline
E-K     & 81.79       &  \textcolor{red}{81.84}      & 82.74       &  \textcolor{red}{83.59}      & 76.29      &  \textcolor{red}{82.84}      & 81.64       &  \textcolor{red}{81.79}      & 81.29       &  \textcolor{red}{81.39}      & 81.24        &  \textcolor{red}{84.94}       & 84.79       &  \textcolor{red}{85.04}       & 82.79       &  \textcolor{red}{83.04}       \\ \hline
K-B     & 73.45       &  \textcolor{red}{73.45}      & 75.60       &  \textcolor{red}{76.35}      & 64.35      &  \textcolor{red}{71.80}      &  \textcolor{red}{71.85}       & 71.45      & 73.05       &  \textcolor{red}{73.85}      & 71.15        &  \textcolor{red}{76.25}       &  \textcolor{red}{75.75}       & 75.70       & 73.50       &  \textcolor{red}{74.20}       \\ \hline
K-D     &  \textcolor{red}{74.44}       & 74.39      & 73.89       &  \textcolor{red}{76.14}      & 73.34      &  \textcolor{red}{74.79}      &  \textcolor{red}{72.99}       & 72.54      & 73.19       &  \textcolor{red}{73.99}      & 71.79        &  \textcolor{red}{76.09}       &  \textcolor{red}{74.39}       & 74.34       & 74.29       &  \textcolor{red}{75.29}       \\ \hline
K-E     & 80.68       &  \textcolor{red}{81.23}      & 81.68       &  \textcolor{red}{82.13}      & 78.63      &  \textcolor{red}{81.68}      & 77.23       &  \textcolor{red}{77.38}      & 79.18       &  \textcolor{red}{79.63}      & 81.23        &  \textcolor{red}{83.33}       &  82.33       & \textcolor{red}{83.18}       & 81.03       &  \textcolor{red}{81.33}       \\ \hline
Mean & 77.16       & \textcolor{red}{77.32}      & 77.80       & \textcolor{red}{79.03}      & 68.14      & \textcolor{red}{76.55}      & 77.05       & \textcolor{red}{77.08}      & 77.17       & \textcolor{red}{77.46}      & 73.85        & \textcolor{red}{79.22}       & 78.07       & \textcolor{red}{78.36}       & 76.99       & \textcolor{red}{77.38}       \\ \hline\hline
Mean4 &67.63	  &\textcolor{red}{68.20}	&64.40	&\textcolor{red}{69.45}	&53.82	&\textcolor{red}{65.21}	&68.09	&\textcolor{red}{68.23}	&68.66	&\textcolor{red}{69.23}	&60.71	&\textcolor{red}{69.97}	&67.48	&\textcolor{red}{69.46}	&68.37	&\textcolor{red}{68.96} \\ \hline
\end{tabular}}
\end{table*}

Herein, we directly regard the artificial data
as the data with new feature representation, i.e., $\mathbf{Z}^\mathcal{S}$ and $\mathbf{Z}^\mathcal{T}$, since they model diverse suboptimal \emph{domain-invariant} alternatives with the domain divergence ranging from low to high.
Note that the dataset with $\mu$ as 0 can be taken as the truly \emph{domain-invariant} case.
We then train two transfer models.
The first model, i.e., $\mathcal{M}_z$, is trained using $\mathbf{Z}^\mathcal{S}$ directly.
The second one is $\widetilde{\mathcal{M}}_z$, specifically RTM$_{lr}^d$.
As the optimal $p$ is different for transfer task with different $\mu$, we compare the best performance across tasks.
Herein, we allow $p$ to take values from 0.05 to 0.95 with 0.05 as step.

The comparison results are shown in Figure \ref{artificial}.
It can be seen that the accuracies of both methods decrease with the increase of $\mu$.
This is reasonable as the data are becoming less and less domain-invariant when $\mu$ grows larger and larger.
Compared $\widetilde{\mathcal{M}}_z$ with $\mathcal{M}_z$, we find that $\widetilde{\mathcal{M}}_z$ consistently achieves better result than $\mathcal{M}_z$.
More specifically, $\widetilde{\mathcal{M}}_z$ yields more significant improvements over $\mathcal{M}_z$ (up to 4.5\% improvements when $\mu = 3$) in the tasks with smaller $\mu$ than the tasks with larger $\mu$.
This is because the more similar the two domains are, the less nuances the
two domains have and the easier RTM$_{lr}^d$ is to achieve the better performance.

\subsection{RTM$_{lr}^d$ on New Data} \label{S5-2}
In this section, we compare $\widetilde{\mathcal{M}}_z$ with $\mathcal{M}_z$ using the new data learned by various state-of-the-art feature-based transfer methods.
The evaluation is done on 4 real-world datasets, namely Office-Caltech10 \cite{gong2012geodesic}, COIL1-COIL2 \cite{long2014transfer}, 20-Newsgroups \cite{Wei2016DeepNF} and Amazon product review dataset \cite{blitzer2007biographies}.

\noindent \textbf{Office-Caltech10 dataset} \cite{gong2012geodesic} is a popular image classification benchmark dataset for transfer learning.
It consists of 4 domains: C (Caltech-256), A (Amazon), W (Webcam), and D (DSLR), where each domain contains images from a specific source.
These 4 domains share the same 10 objects, but have different data distributions.
In our experiment, we construct 12 transfer tasks, denoted as `source-target'.
For instance, C-A means that C is the source domain
and A is the target domain.
Following \cite{gong2012geodesic}, We use SURF features.

\noindent \textbf{COIL1-COIL2 dataset} \cite{long2014transfer} is another image dataset for object recognition.
Two domains, COIL1 and COIL2, share 20 objects.
Each of them has 720 images, and each image is $32\times32$ pixels with 256 gray levels per pixel.
The images in two domains are taken in different directions, and thus are drawn from different distributions.
COIL1 contains images taken in the directions of $[0^\circ,85^\circ]\cup[180^\circ,265^\circ]$.
COIL2 contains images taken in the directions of $[90^\circ,175^\circ]\cup[270^\circ,355^\circ]$.
We construct two tasks: CL$_{12}$ and CL$_{21}$.

\noindent \textbf{20-Newsgroups Dataset} \cite{Wei2016DeepNF} consists of about 20,000 documents from 4 top categories: computer (C), recording (R), science (S), and talk (T). Each top category has 4 subcategories.
Top categories are treated as labels, while subcategories are treated as related domains.
Six binary prediction tasks are formed: C-R, C-S, C-T, R-S, R-T, and S-T. We take the task C-R for instance.
Top category C is the positive class and R is the negative class.
Two subcategories under each class are selected to constitute the source domain, while another two subcategories are selected to form the target domain.
By exchanging the roles of the two domains, we have two domain pairs for the prediction task C-R, denoted as CR-1 and CR-2.

\noindent \textbf{Amazon product review dataset} \cite{blitzer2007biographies} contains sentiment reviews from 4 product categories: books (B), DVD (D), electronics (E) and kitchen appliance (K). Each review is characterized by unigram and bigram tf-idf features and labeled as positive or negative. Each domain has about 2,000 samples. When a domain is selected as source/target, all the samples in this domain are used as training/test data. By pairing up the domains for adaptation task, we have 12 domain pairs, denoted as `source-target'.
For example, B-D means that category B is the source domain and D is the target domain.

\subsubsection{Results on New Data of Subspace-based Methods} \label{S5-2-1}
For each dataset, we first adopt existing subspace-based transfer methods to learn the new data, including TCA \cite{pan2011domain}, GFK \cite{gong2012geodesic}, SA \cite{fernando2013unsupervised}, TJM \cite{long2014transfer}, JDA \cite{long2013transfer}, CORAL \cite{sun2016return}, JGSA \cite{zhang2017joint}, and MMIT \cite{wei2019knowledge}.
For all the methods, we uniformly set the subspace dimensionality as 20.
Regarding the hyper-parameters in each method, we follow the default ones specified by the authors.
As the aim is to show the superiority of RTM$_{lr}^d$ to the conventional transfer model, we compare the best results achieved by $\widetilde{\mathcal{M}}_z$ and $\mathcal{M}_z$.
We use the best performance with the optimal $p$ in each task where $p$ takes values from 0.05 to 0.95 with 0.05 as step.
For fair comparison, we also use linear regression model for $\mathcal{M}_z$\footnote{Note that most of existing methods use Nearest Neighbors as the base classifier, which makes the reported results in this paper different from those reported in existing papers.}.
All the comparison results are shown in Table \ref{compaful}.
For each transfer task, we highlight the winner between $\mathcal{M}_z$ and $\widetilde{\mathcal{M}}_z$ using red.
We also show the average results of all the transfer tasks in each datasets, and the average results of all the four datasets.

Regarding the average result of all the transfer tasks, $\widetilde{\mathcal{M}}_z$ consistently outperforms $\mathcal{M}_z$ in all different methods. The improvement is up to 11.39\% in SA, and the mean of the improvement on the 8 methods is 3.69\%. This demonstrates the superiority of $\widetilde{\mathcal{M}}_z$  over $\mathcal{M}_z$.
Regarding the average result of each dataset, we find that $\widetilde{\mathcal{M}}_z$ sweeps $\mathcal{M}_z$ in 3 of 4 datasets. The only exception is TJM in 20-Newsgroups dataset, where $\widetilde{\mathcal{M}}_z$ achieves slightly worse result, 0.15\% less, than $\mathcal{M}_z$. The means of the improvement on the 8 methods are  7.59\%, 4.64\%, 3.77\%, 2.02\% for COIL1-COIL2, Office-Caltech10, 20-Newsgroup, and Amazon product review dataset, respectively.
These results verify the effectiveness of RTM$_{lr}^d$ in diverse transfer tasks as well as on diverse new data.


\subsubsection{Results on New Data of Deep Transfer Methods} \label{S5-2-2}
We further compare $\widetilde{\mathcal{M}}_z$ and $\mathcal{M}_z$ on the new data learned from existing deep learning based transfer methods, including mSDA \cite{chen2012marginalized}, DNFC \cite{Wei2016DeepNF}, TarReg \cite{clinchant2016domain}.
These 3 deep methods only focus on unsupervised feature learning, that is, they decouple the feature learning with the transfer model training into two separate stages\footnote{We do not use the end-to-end deep learning methods, e.g., DAN \cite{long2015learning}, DANN \cite{ganin2016domain}, and JAN \cite{tzeng2017adversarial}, to generate new data to avoid the extremely expensive computational cost.
One needs to run the whole deep learning procedure and then extract out the new data, which is very time-consuming and computationally expensive.}.
The results are shown in Table \ref{COMPDEEP}.
\begin{table}[t]
\scriptsize
\caption{Comparisons of $\widetilde{\mathcal{M}}_z$ and $\mathcal{M}_z$ on the feature representation of deep learning based transfer methods.}
    \label{COMPDEEP}
\begin{minipage}{1\linewidth}
\centering
\begin{tabular}{|c|c|c|c|c|c|c|}
\hline
  Office     & \multicolumn{2}{c|}{mSDA} & \multicolumn{2}{c|}{DNFC} & \multicolumn{2}{c|}{TarReg} \\ \hline
  Method     & $\mathcal{M}_z$        & $\widetilde{\mathcal{M}}_z$         & $\mathcal{M}_z$        & $\widetilde{\mathcal{M}}_z$         & $\mathcal{M}_z$        & $\widetilde{\mathcal{M}}_z$          \\ \hline \hline
C-A    & 40.40       & \textcolor{red}{55.95}       & 31.11       & \textcolor{red}{56.68}       & \textcolor{red}{55.43}        & 51.88        \\ \hline
C-W    & 32.20       & \textcolor{red}{56.27}       & 24.41       & \textcolor{red}{51.19}       & 44.41        & \textcolor{red}{51.53}        \\ \hline
C-D    & 30.57       & \textcolor{red}{55.41}       & 29.30       & \textcolor{red}{50.32}       & 44.59        & \textcolor{red}{54.14}        \\ \hline
A-C    & 37.22       & \textcolor{red}{46.84}       & 25.11       & \textcolor{red}{46.48}       & \textcolor{red}{43.37}        & 43.10        \\ \hline
A-W    & 32.20       & \textcolor{red}{44.75}       & 27.46       & \textcolor{red}{44.75}       & 36.61        & \textcolor{red}{43.05}        \\ \hline
A-D    & 33.12       & \textcolor{red}{45.86}       & 26.75       & \textcolor{red}{44.59}       & 40.13        & \textcolor{red}{44.59}        \\ \hline
W-C    & 35.08       & \textcolor{red}{37.93}       & 33.84       & \textcolor{red}{37.40}       & 35.35        & \textcolor{red}{37.85}        \\ \hline
W-A    & 36.95       & \textcolor{red}{42.17}       & 37.79       & \textcolor{red}{42.69}       & 36.01        & \textcolor{red}{41.54}        \\ \hline
W-D    & 86.62       & \textcolor{red}{89.17}       & 84.08       & \textcolor{red}{89.17}       & 78.34        & \textcolor{red}{78.34}        \\ \hline
D-C    & 34.73       & \textcolor{red}{38.11}       & 32.86       & \textcolor{red}{36.42}       & 24.22        & \textcolor{red}{39.63}        \\ \hline
D-A    & 38.62       & \textcolor{red}{43.01}       & 33.92       & \textcolor{red}{39.35}       & 26.62        & \textcolor{red}{41.96}        \\ \hline
D-W    & 85.42       & \textcolor{red}{87.12}       & 77.97       & \textcolor{red}{86.10}       & 61.36        & \textcolor{red}{85.08}        \\ \hline
Mean   & 43.60       & \textcolor{red}{53.55}       & 38.72       & \textcolor{red}{52.09}       & 43.87        & \textcolor{red}{51.06}        \\ \hline
    \end{tabular}
  \end{minipage}

   \vspace{5pt}
  \begin{minipage}{1\linewidth}
    \centering
    \begin{tabular}{|c|c|c|c|c|c|c|}
\hline
   20News    & \multicolumn{2}{c|}{mSDA} & \multicolumn{2}{c|}{DNFC} & \multicolumn{2}{c|}{TarReg} \\ \hline
   Method    & $\mathcal{M}_z$        & $\widetilde{\mathcal{M}}_z$         & $\mathcal{M}_z$        & $\widetilde{\mathcal{M}}_z$         & $\mathcal{M}_z$        & $\widetilde{\mathcal{M}}_z$          \\ \hline \hline
       C-R    & 68.72       & \textcolor{red}{75.49}       & 68.47       & \textcolor{red}{72.32}       & 78.15        & \textcolor{red}{85.95}        \\ \hline
C-S    & 84.35       & \textcolor{red}{88.54}       & 77.25       & \textcolor{red}{81.52}       & 81.52        & \textcolor{red}{91.10}        \\ \hline
C-T    & 85.36       & \textcolor{red}{90.62}       & 81.82       & \textcolor{red}{85.36}       & 90.62        & \textcolor{red}{94.74}        \\ \hline
R-C    & 81.51       & \textcolor{red}{90.08}       & 79.73       & \textcolor{red}{83.80}       & 88.13        & \textcolor{red}{94.15}        \\ \hline
R-S    & 76.45       & \textcolor{red}{79.23}       & 73.00       & \textcolor{red}{75.36}       & 81.50        & \textcolor{red}{90.66}        \\ \hline
R-T    & \textcolor{red}{71.40}       & 64.20       & 71.69       & \textcolor{red}{72.25}       & 80.40        & \textcolor{red}{84.19}        \\ \hline
S-C    & 78.02       & \textcolor{red}{82.28}       & 72.74       & \textcolor{red}{74.36}       & 78.53        & \textcolor{red}{87.14}        \\ \hline
S-R    & 84.65       & \textcolor{red}{89.80}       & 76.31       & \textcolor{red}{78.41}       & 87.18        & \textcolor{red}{94.27}        \\ \hline
S-T    & \textcolor{red}{80.04}       & 76.25       & 66.79       & \textcolor{red}{74.17}       & 81.65        & \textcolor{red}{87.89}        \\ \hline
T-C    & 89.27       & \textcolor{red}{91.17}       & 85.07       & \textcolor{red}{87.38}       & 93.06        & \textcolor{red}{96.53}        \\ \hline
T-R    & 56.26       & \textcolor{red}{69.10}       & 77.52       & \textcolor{red}{81.83}       & 80.60        & \textcolor{red}{94.66}        \\ \hline
T-S    & 81.39       & \textcolor{red}{86.97}       & 79.01       & \textcolor{red}{82.42}       & 87.59        & \textcolor{red}{94.52}        \\ \hline
Mean   & 78.12       & \textcolor{red}{81.98}       & 75.78       & \textcolor{red}{79.10}       & 84.08        & \textcolor{red}{91.32}        \\ \hline
    \end{tabular}
  \end{minipage}

  \vspace{5pt}
    \begin{minipage}{1\linewidth}
    \centering
    \begin{tabular}{|c|c|c|c|c|c|c|}
\hline
   Amazon    & \multicolumn{2}{c|}{mSDA} & \multicolumn{2}{c|}{DNFC} & \multicolumn{2}{c|}{TarReg} \\ \hline
   Method    & $\mathcal{M}_z$        & $\widetilde{\mathcal{M}}_z$         & $\mathcal{M}_z$        & $\widetilde{\mathcal{M}}_z$         & $\mathcal{M}_z$        & $\widetilde{\mathcal{M}}_z$          \\ \hline \hline
       B-D    & 68.93       & \textcolor{red}{81.49}       & 76.09       & \textcolor{red}{80.74}       & 78.84        & \textcolor{red}{80.04}        \\ \hline
B-E    & 61.71       & \textcolor{red}{81.78}       & 71.97       & \textcolor{red}{79.88}       & 74.52        & \textcolor{red}{80.78}        \\ \hline
B-K    & 65.58       & \textcolor{red}{82.29}       & 72.19       & \textcolor{red}{81.09}       & 76.59        & \textcolor{red}{82.19}        \\ \hline
D-B    & 68.90       & \textcolor{red}{80.65}       & 65.65       & \textcolor{red}{78.00}       & 78.40        & \textcolor{red}{80.05}        \\ \hline
D-E    & 61.16       & \textcolor{red}{81.08}       & 60.01       & \textcolor{red}{79.48}       & 74.52        & \textcolor{red}{82.13}        \\ \hline
D-K    & 66.48       & \textcolor{red}{82.64}       & 69.73       & \textcolor{red}{81.34}       & 76.79        & \textcolor{red}{82.99}        \\ \hline
E-B    & 64.95       & \textcolor{red}{77.20}       & 66.40       & \textcolor{red}{75.20}       & 73.50        & \textcolor{red}{76.65}        \\ \hline
E-D    & 64.13       & \textcolor{red}{77.74}       & 70.89       & \textcolor{red}{75.84}       & 73.24        & \textcolor{red}{77.59}        \\ \hline
E-K    & 74.74       & \textcolor{red}{85.99}       & 76.64       & \textcolor{red}{83.94}       & 84.14        & \textcolor{red}{85.54}        \\ \hline
K-B    & 68.30       & \textcolor{red}{77.60}       & 67.10       & \textcolor{red}{75.90}       & 73.65        & \textcolor{red}{77.35}        \\ \hline
K-D    & 66.23       & \textcolor{red}{77.74}       & 69.08       & \textcolor{red}{76.64}       & 73.49        & \textcolor{red}{76.74}        \\ \hline
K-E    & 77.43       & \textcolor{red}{85.14}       & 80.73       & \textcolor{red}{84.48}       & 82.88        & \textcolor{red}{83.33}        \\ \hline
Mean   & 67.38       & \textcolor{red}{80.95}       & 70.54       & \textcolor{red}{79.38}       & 76.71        & \textcolor{red}{80.45}        \\ \hline
    \end{tabular}
  \end{minipage}

  \vspace{5pt}
      \begin{minipage}{1\linewidth}
    \centering
    \begin{tabular}{|c|c|c|c|c|c|c|}
\hline
   COIL    & \multicolumn{2}{c|}{mSDA} & \multicolumn{2}{c|}{DNFC} & \multicolumn{2}{c|}{TarReg} \\ \hline
   Method    & $\mathcal{M}_z$        & $\widetilde{\mathcal{M}}_z$         & $\mathcal{M}_z$        & $\widetilde{\mathcal{M}}_z$         & $\mathcal{M}_z$        & $\widetilde{\mathcal{M}}_z$          \\ \hline \hline
       CL$_{12}$ & 79.03       & \textcolor{red}{83.06}       & 81.94       & \textcolor{red}{82.22}       & 69.58        & \textcolor{red}{71.25}        \\ \hline
CL$_{21}$ & 79.17       & \textcolor{red}{81.67}       & \textcolor{red}{82.64}       & 82.50       & \textcolor{red}{69.86}        & 69.03        \\ \hline
Mean   & 79.10       & \textcolor{red}{82.36}       & 82.29       & \textcolor{red}{82.36}       & 69.72        & \textcolor{red}{70.14}        \\ \hline \hline
Mean4  & 63.88       & \textcolor{red}{72.69}       & 62.76       & \textcolor{red}{70.83}       & 68.30        & \textcolor{red}{74.06}        \\ \hline
    \end{tabular}
  \end{minipage}
\end{table}

\begin{figure*}[t]
\centering
\subfloat
{\includegraphics[scale = 0.4]{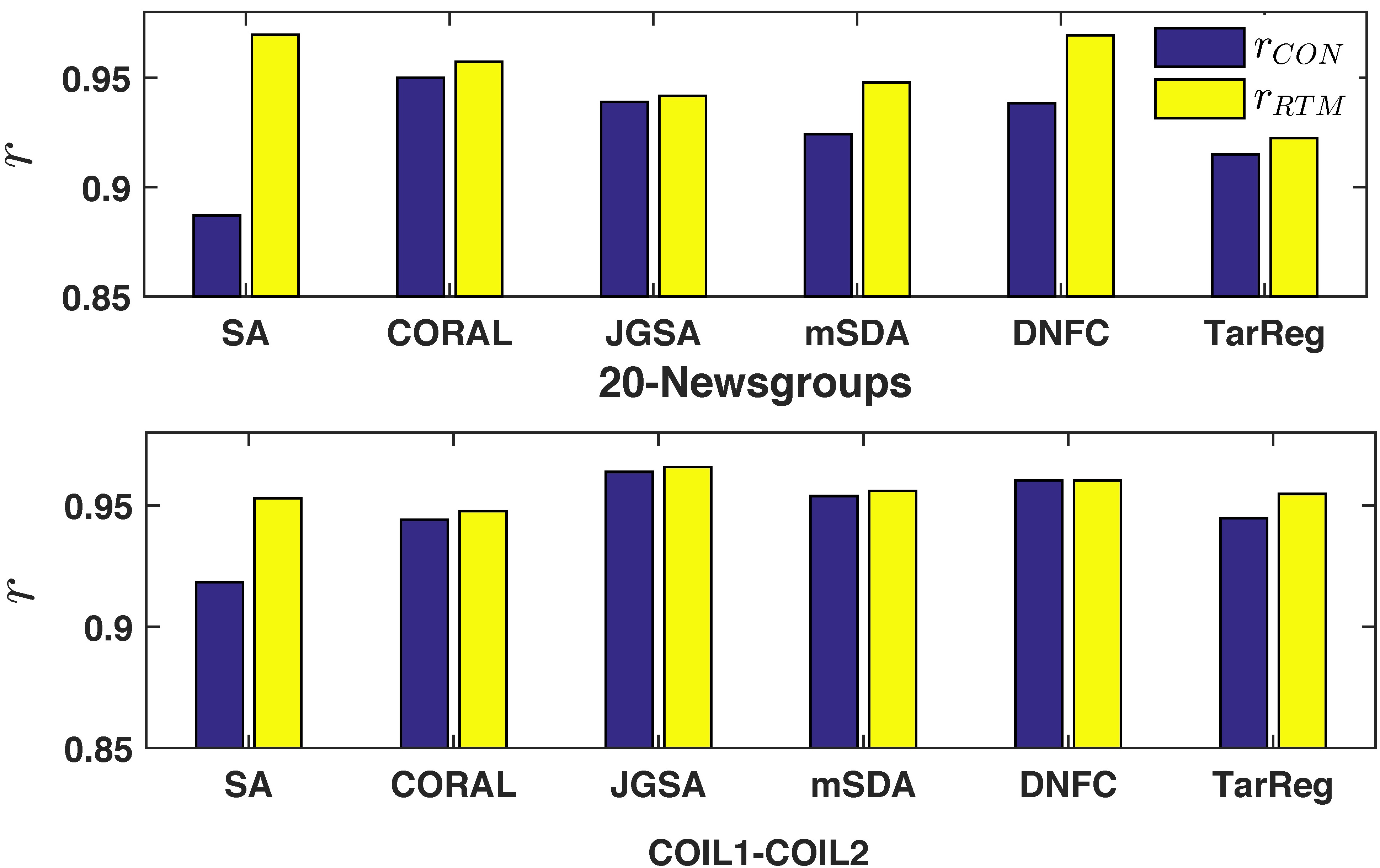}}
\subfloat
{\includegraphics[scale = 0.39]{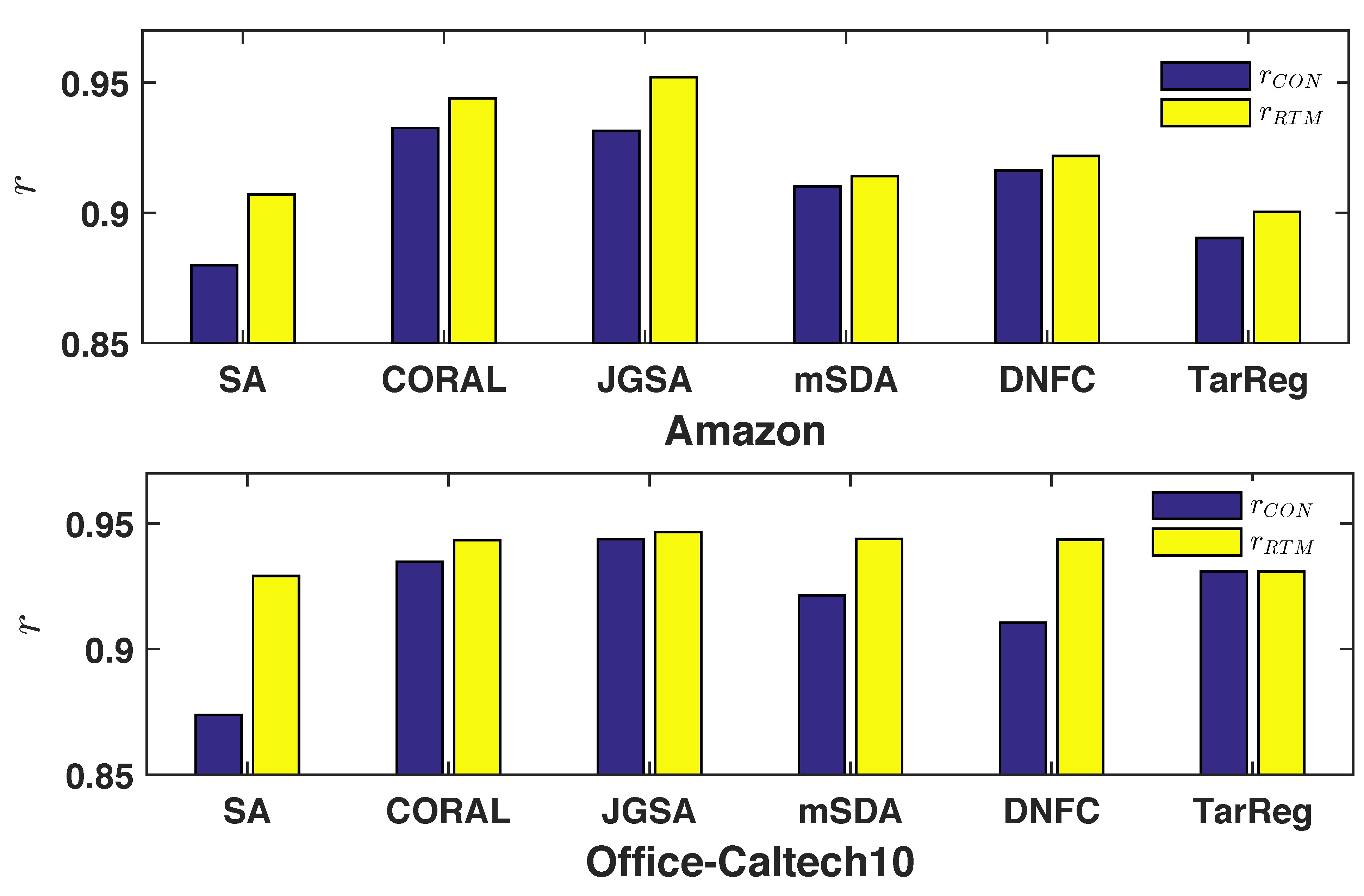}}
 \caption{NDCG Results on the 4 datasets.
 } \label{ndcg}
\end{figure*}
\begin{figure*}[t]
\centering
\resizebox{\linewidth}{!}{
\subfloat [Results on Amazon dataset.]
{\includegraphics[scale = 0.34]{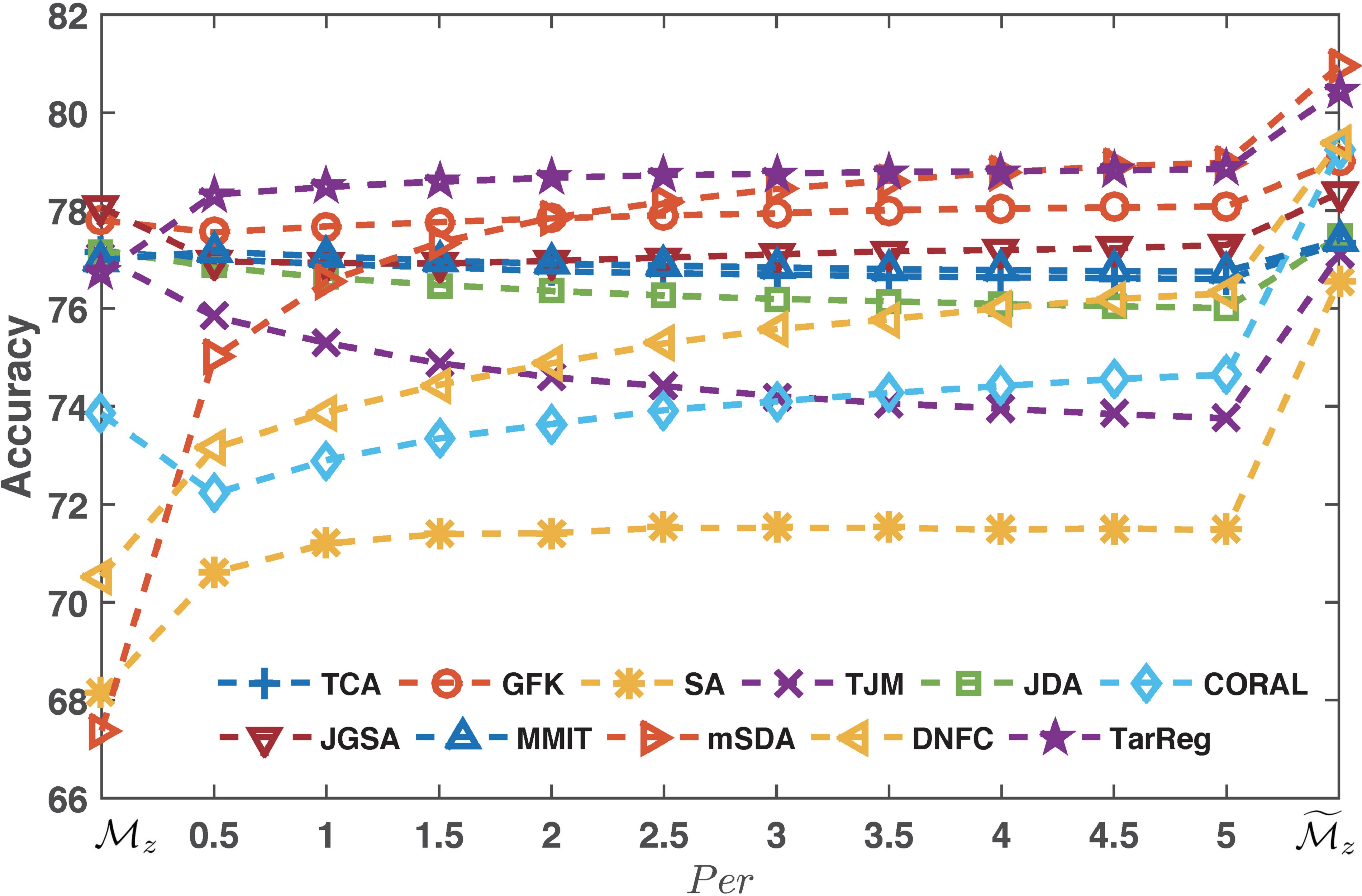}\label{amazonresult}}
\subfloat [Results on 20-newsgroups dataset.]
{\includegraphics[scale = 0.34]{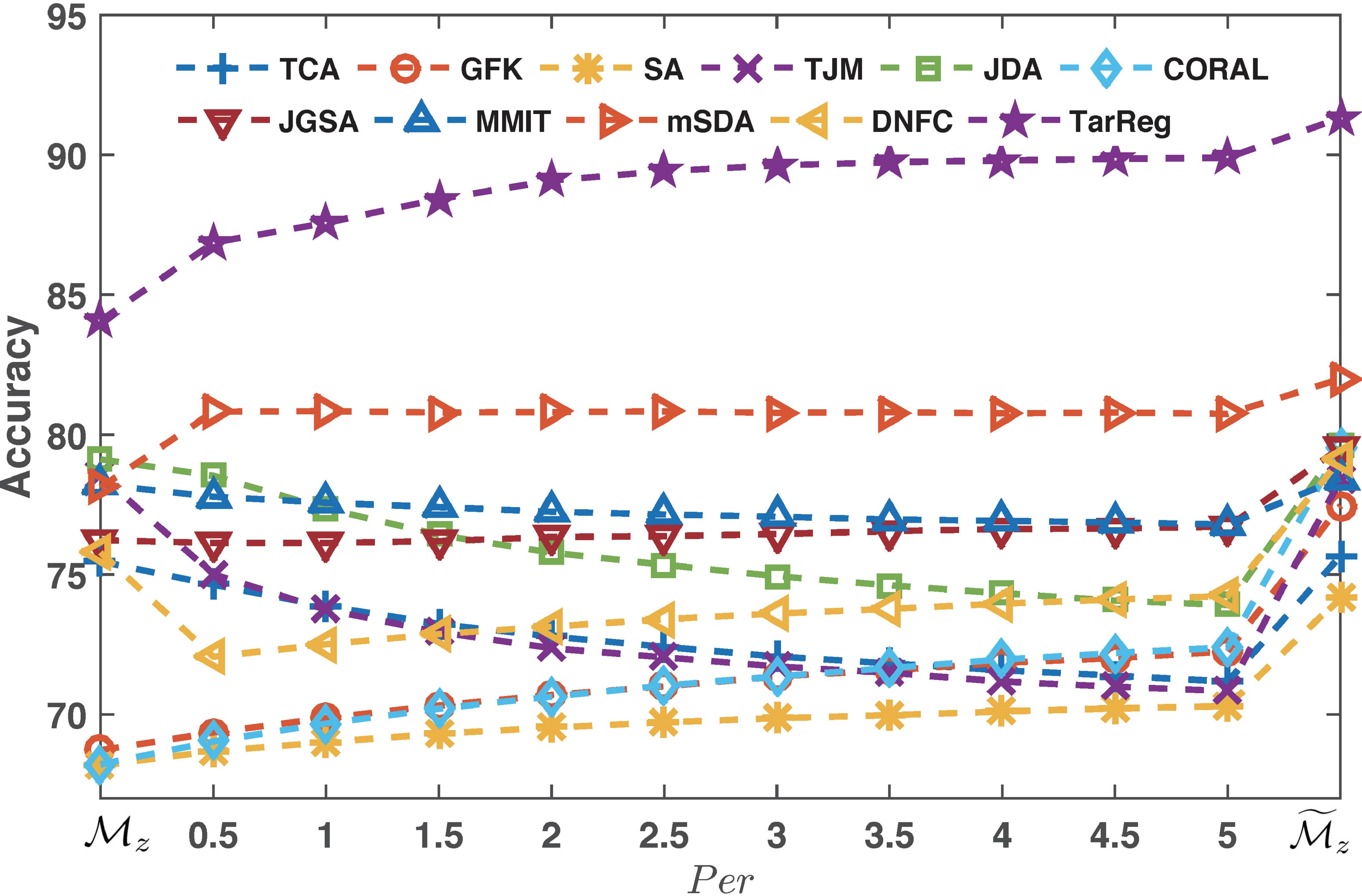}}
\subfloat [Results on COIL1-COIL2 dataset.]
{\includegraphics[scale = 0.34]{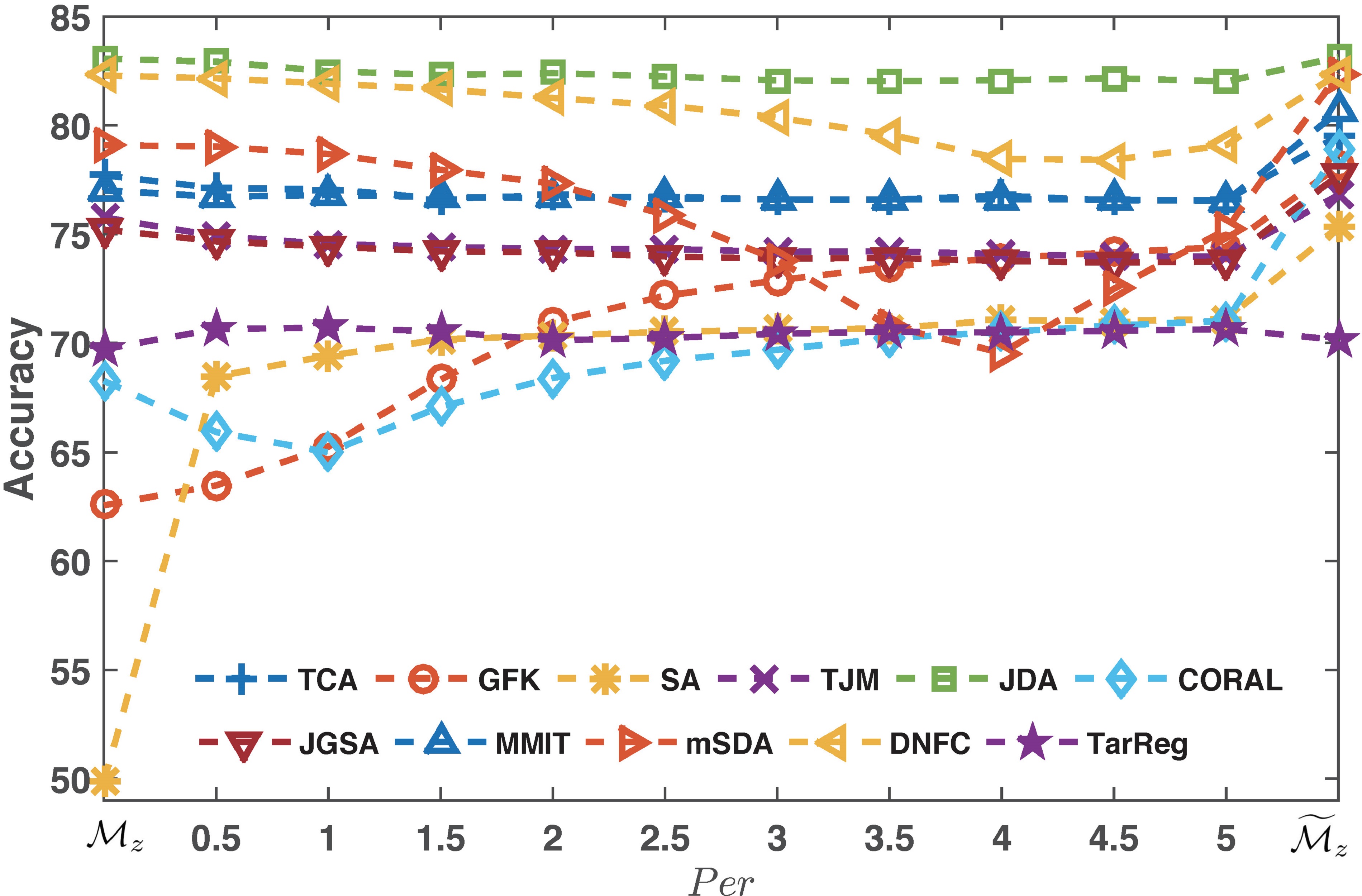}}
\subfloat [Results on OfficeCaltech10 dataset.]
{\includegraphics[scale = 0.34]{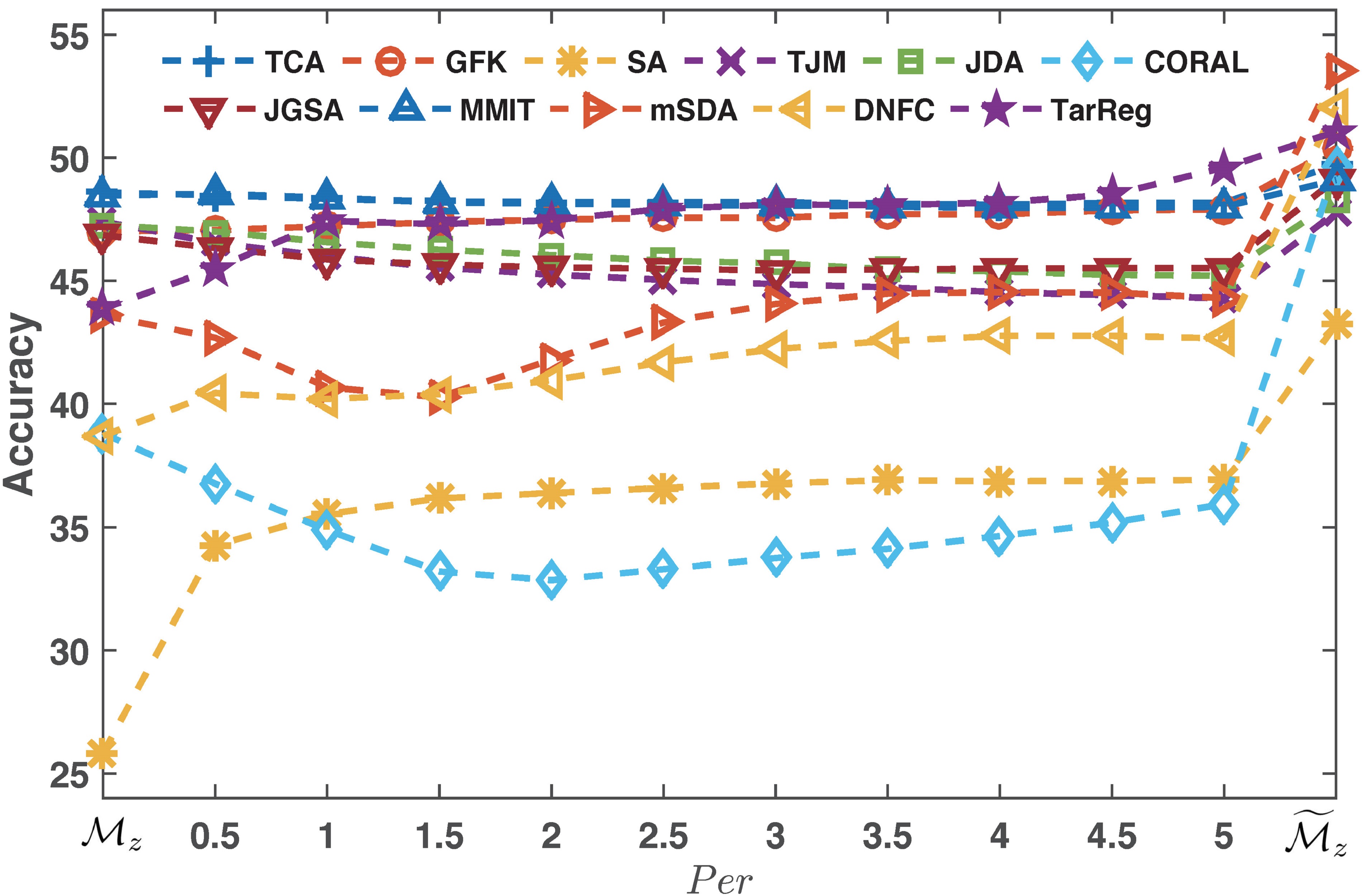}}}
 \caption{Data random augmentation results on the 4 datasets.
 } \label{RanAug}
\end{figure*}

From Table \ref{COMPDEEP}, we observe that, overall, $\widetilde{\mathcal{M}}_z$ consistently achieves better results than $\mathcal{M}_z$ regarding the average result of each dataset as well as the average result of all the 4 datasets.
This again verifies the superiority of $\widetilde{\mathcal{M}}_z$ to $\mathcal{M}_z$.
Furthermore, we find that (1) $\widetilde{\mathcal{M}}_z$ in this deep learning based experiment basically yields much better results than that in the subspace-based experiment, e.g., the winner $\widetilde{\mathcal{M}}_z$ of these two sets of experiments is CORAL (69.97\%) and TarReg (74.06\%), and TarReg outperforms CORAL with 4.09\% improvements.
(2) The improvements of $\widetilde{\mathcal{M}}_z$ over $\mathcal{M}_z$ in the deep learning based experiment are generally more significant than those in subspace-based experiment.
The average improvement on the 8 subspace-based methods is 3.69\%, while the average improvement on the 3 deep feature based methods is 7.55\%.
Such performance discrepancy between the two sets of experiments may be due to the feature dimensionality of the new data.
Deep feature based methods discover several deep-layer features and enlarge the feature dimensionality.
This enables richer and more diverse corruptions modelled by RTM$_{lr}^d$.
However, subspace-based methods reduce the feature dimensionality, and thus the feature corruptions modelled by RTM$_{lr}^d$ are much less than that of deep feature based methods.

\subsubsection{Model Similarity Analysis}
To further show the superiority of $\widetilde{\mathcal{M}}_z$ to ${\mathcal{M}}_z$, it is desired to visualize how the randomly corrupted source data are distributed so that we can clearly see whether they cover the new target data or not.
However, it is infeasible as RTM actually does not conduct any corruption.
This instead drives us to directly pay attention to the hyperplanes learned by $\widetilde{\mathcal{M}}_z$ and ${\mathcal{M}}_z$.
In this section, we aim to analyze the similarity of the model weight importance of a transfer model with that of the target ground-truth one.
Specifically, we obtain a ground-truth target hyperplane by training a linear regression model using target labelled data, and then we want to know which hyperplane of $\widetilde{\mathcal{M}}_z$ and ${\mathcal{M}}_z$ is closer to the target one in terms of model weight importance.
We use the normalized discounted cumulative gain (NDCG) \cite{jarvelin2002cumulated} to measure the similarity of two ranks.
Within our context, we have:
\begin{equation} \nonumber
r = \frac{NDCG_{\mathcal{S}}}{NDCG_{\mathcal{T}}}.
\end{equation}
The closer to 1 $r$ is, the closer the feature importance of the source and target is.
We then calculate $r_{CON}$ and $r_{RTM}$ by using model weights from ${\mathcal{M}}_z$ and $\widetilde{\mathcal{M}}_z$, respectively.
We test $\widetilde{\mathcal{M}}_z$ on SA, CORAL, and JGSA as they are the top 3 methods regarding the improvements achieved from ${\mathcal{M}}_z$ to $\widetilde{\mathcal{M}}_z$ in Table \ref{compaful}.
We also test on the 3 deep feature based methods.
For each method, we report the average $r$ of all the tasks given a dataset.
The results on the 4 datasets are shown in Figure \ref{ndcg}.
We can see that $r_{RTM}$ is consistently larger than $r_{CON}$, which indicates that the model weight importance of $\widetilde{\mathcal{M}}_z$ is much more similar with the target ground-truth one than that of ${\mathcal{M}}_z$.
This explains why $\widetilde{\mathcal{M}}_z$ can achieve better transfer performance than ${\mathcal{M}}_z$.

\subsubsection{Random Data Augmentation Analysis} \label{S5-2-4}
Using the marginalization trick, RTM$_{lr}^d$ is equivalent to being trained on noisy source data populations with infinite corruptions.
As RTM$_{lr}^d$ does not actually conduct any corruption, we are curious about the performance of a model variant $\overline{\mathcal{M}}_z$ that is trained using actual random data augmentations.
To do so, we augment the new source data by conducting random corruptions.
For fair comparison, we also use dropout noise.
Specifically, we independently and uniformly sample $per \times n$ data points from the new source data where $n$ is the data size and $per$ controls how much corruptions will be made.
We then independently corrupt each sampled data using dropout noise where each feature dimension of the sampled data would be corrupted to be 0 with a probability $p$.
We let $p$ take 10 values from 0.1 to 0.9 with 0.1 as step to diversify random corruptions.
The corrupted data are assigned with the same label as the data point from which it is corrupted.
To investigate the effect of the number of corruptions on the model performance, we allow $per$ to take values from 0.5 to 5 with 0.5 as step.
Then we have 10 sets of data augmentations with the augmentation size from small to large.
To be statistically stable, we run 20 times for each of the 10 sets.
Each set equivalently conducts $10\times20\times per\times n$ total data augmentations.
For each set, we report the average result of the corresponding entire augmentations.
We also plot the results of $\mathcal{M}_z$ and $\widetilde{\mathcal{M}}_z$ at the first and last column of the figure for comparisons.
We test $\overline{\mathcal{M}}_z$ trained using new data learned from 8 subspace-based methods and 3 deep feature based methods.

Herein, we show the results on the 4 datasets in Figure \ref{RanAug}.
From Figure \ref{RanAug}, we obtain the following observations.
Firstly, compared with $\mathcal{M}_z$, $\overline{\mathcal{M}}_z$ trained using the new data from different methods performs differently.
For instance, $\overline{\mathcal{M}}_z$ with SA features (yellow `$\ast$') further boosts the performance of $\mathcal{M}_z$ while $\overline{\mathcal{M}}_z$ with TJM features (purple `$\times$') generally fails to do so.
This shows that random data augmentation is not guaranteed to enhance model performance.
Secondly, we observe that as the number of augmentations increases, the trend of the performance variation of $\overline{\mathcal{M}}_z$ differs considerably across different methods.
For instance in Figure \ref{amazonresult}, for DNFC $\overline{\mathcal{M}}_z$ benefits from the increasing augmentations, while for GFK and TarReg, it is either unaffected or negatively affected.
However, a clear improvement of $\widetilde{\mathcal{M}}_z$ over $\overline{\mathcal{M}}_z$ is observed at the last column for all the methods.
This is because RTM$_{lr}^d$ simulates the infinite case, and thus are more powerful.
These experiments show that RTM$_{lr}^d$ is more effective than models trained with actual random data augmentation.

Moreover, we want to highlight that the model performance of $\overline{\mathcal{M}}_z$ is very unstable.
Firstly, for a specific feature representation, the model performance variation trend with increasing augmentations differ considerably across datasets, e.g., for mSDA features, flat on 20-Newsgroups dataset, but decreasing first and then increasing on COIL1-COIL2 dataset.
A more extreme case is DNFC, that yields completely opposite performance between COIL1-COIL2 dataset (decreasing) and OfficeCaltech10 dataset (increasing).
Secondly, for a specific dataset, the model performance variation trends also have a big difference across methods, e.g., SA and CORAL in OfficeCaltech10 dataset.
All the observations show us $\overline{\mathcal{M}}_z$ is not a good stable transfer model.

\subsection{RTM$_{lr}^d$ on the Original Data} \label{S5-3}
In this section, we investigate the performance of RTM$_{lr}^d$ on the original data.
Specifically, we compare $\widetilde{\mathcal{M}}_x$ with $\mathcal{M}_x$ on 5 real-world datasets.
We select 2 benchmark datasets widely used for subspace-based transfer methods, i.e., Office-Caltech10 and 20-Newsgroups, and we further use 3 benchmark datasets that is widely used in the evaluations of the deep learning transfer methods.
The ResNet-50 features are used.

\noindent \textbf{Office-31} \cite{long2015learning} is a benchmark dataset for visual domain adaptation. It contains 4,652 images and 31 categories from 3 distinct domains: Amazon (A), Webcam (W) and DSLR (D). By pairing up 2 domains, we construct 6 transfer tasks.

\noindent \textbf{ImageCLEF-DA} \cite{Long2017Deep} is built for ImageCLEF 2014 domain adaptation challenge. It contains 4 domains including Caltech-256 (C), ImageNet ILSVRC 2012 (I), Bing (B) and Pascal VOC 2012 (P). Following \cite{Long2017Deep}, we construct 6 transfer tasks.

\noindent \textbf{Office-Home} \cite{cai2019learning} is a more challenging domain adaptation dataset, which consists of around 15,500 images from 65 categories of everyday objects. This dataset is organized into 4 domains: Art (Ar), Clipart (Cl), Product (Pr) and Real-world (Rw). By pairing up 2 domains, we have 12 transfer tasks.
\begin{table}[t]
\centering
\caption{$\widetilde{\mathcal{M}}_x$ and $\mathcal{M}_x$ for subspace case. }
    \label{subsapce}
\renewcommand\arraystretch{1}
\begin{tabular}{|c|c|c|c|c|c|}
\hline
\multirow{2}{*}{Method} & \multirow{2}{*}{$\mathcal{M}_x$} & \multicolumn{3}{c|}{$\mathcal{M}_z$}                          & \multirow{2}{*}{$\widetilde{\mathcal{M}}_x$} \\ \cline{3-5}
                        &                     & JDA            & JGSA           & MMIT           &                     \\ \hline \hline
C-A                     & 33.30               & 49.27          & 51.67          & 51.98          & \textbf{55.32}      \\ \hline
C-W                     & 26.44               & 48.14          & 45.42          & 49.15          & \textbf{49.49}      \\ \hline
C-D                     & 25.48               & 47.13          & 40.76          & 47.13          & \textbf{47.13}      \\ \hline
A-C                     & 25.02               & 38.91          & 42.65          & 42.56          & \textbf{44.08}      \\ \hline
A-W                     & 20.00               & \textbf{46.78} & 35.59          & 40.00          & 41.02               \\ \hline
A-D                     & 24.20               & \textbf{48.41} & 37.58          & 41.40          & 42.68               \\ \hline
W-C                     & 31.17               & 32.32          & \textbf{34.55} & 38.11          & 34.46               \\ \hline
W-A                     & 35.18               & 36.12          & 38.31          & \textbf{40.29} & 39.46               \\ \hline
W-D                     & 80.25               & 74.52          & 82.80          & 76.43          & \textbf{83.44}      \\ \hline
D-C                     & 31.97               & 32.50          & 32.95          & \textbf{35.71} & 32.77               \\ \hline
D-A                     & 35.59               & 35.59          & 35.91          & \textbf{39.35} & 36.64               \\ \hline
D-W                     & 80.68               & 77.63          & 84.07          & 79.32          & \textbf{84.75}      \\ \hline
Mean                    & 37.44               & 47.28          & 46.86          & 48.45          & \textbf{49.27}      \\ \hline \hline
CR-1                    & 60.50               & 64.52          & 63.24          & 71.12          & \textbf{69.58}      \\ \hline
CR-2                    & 74.89               & 81.51          & 83.04          & 81.51          & \textbf{84.22}      \\ \hline
CS-1                    & 68.35               & \textbf{87.17} & 77.16          & 78.44          & 78.61               \\ \hline
CS-2                    & 70.19               & 73.25          & 72.74          & 71.29          & \textbf{75.72}      \\ \hline
CT-1                    & 63.54               & 86.12          & 86.89          & 86.22          & \textbf{90.05}      \\ \hline
CT-2                    & 81.91               & 88.12          & 89.70          & 88.12          & \textbf{90.12}      \\ \hline
RS-1                    & 70.06               & \textbf{78.64} & 70.73          & 72.83          & 75.36               \\ \hline
RS-2                    & 72.26               & \textbf{81.45} & 79.09          & 75.13          & 79.93               \\ \hline
RT-1                    & 60.98               & 75.57          & 64.11          & \textbf{77.37} & 68.84               \\ \hline
RT-2                    & 69.82               & \textbf{81.11} & 67.15          & 80.80          & 75.98               \\ \hline
ST-1                    & 58.85               & 69.06          & 77.29          & 73.60          & \textbf{79.85}      \\ \hline
ST-2                    & 67.11               & 82.94          & 83.66          & 82.21          & \textbf{86.14}      \\ \hline
Mean                    & 68.21               & 79.12          & 76.23          & 78.22          & \textbf{79.53}      \\ \hline \hline
Mean2                   & 52.82               & 63.20          & 61.54          & 63.34          & \textbf{64.40}      \\ \hline
\end{tabular}
\end{table}
\begin{table}[t]
\centering
\caption{$\widetilde{\mathcal{M}}_x$ and $\mathcal{M}_x$ for deep learning case. }
    \label{deepsupp}
\renewcommand\arraystretch{1}
\begin{tabular}{|c|c|c|c|c|c|}
\hline
\multirow{2}{*}{Method} & \multirow{2}{*}{$\mathcal{M}_x$} & \multicolumn{3}{c|}{$\mathcal{M}_z$}              & \multirow{2}{*}{$\widetilde{\mathcal{M}}_x$} \\ \cline{3-5}
                        &                     & DAN  & DANN          & JAN           &                     \\ \hline \hline
A-W                     & 67.7                & 80.5 & 82.0          & \textbf{85.4} & 74.5                \\ \hline
A-D                     & 74.3                & 78.6 & 79.7          & \textbf{84.7} & 80.3                \\ \hline
W-A                     & 58.8                & 62.8 & 67.4          & \textbf{70.0} & 63.2                \\ \hline
W-D                     & 96.2                & 99.6 & 99.1          & \textbf{99.8} & 96.6                \\ \hline
D-A                     & 57.1                & 63.6 & 68.2          & \textbf{68.6} & 61.0                \\ \hline
D-W                     & 91.7                & 97.1 & 96.9          & \textbf{97.4} & 92.6                \\ \hline
Mean                    & 74.3                & 80.4 & 82.2          & \textbf{84.3} & 78.0                \\ \hline \hline
C-I                     & 75.8                & 86.3 & \textbf{87.0} & 86.9          & 82.3                \\ \hline
C-P                     & 60.0                & 69.2 & \textbf{74.3} & 72.7          & 66.0                \\ \hline
I-C                     & 82.0                & 92.8 & \textbf{96.2} & 95.3          & 85.8                \\ \hline
I-P                     & 65.8                & 74.5 & 75.0          & \textbf{75.6} & 70.3                \\ \hline
P-C                     & 75.8                & 89.8 & 91.5          & \textbf{92.2} & 84.3                \\ \hline
P-I                     & 73.0                & 82.2 & 86.0          & \textbf{86.8} & 83.3                \\ \hline
Mean                    & 72.1                & 82.5 & \textbf{85.0} & 84.9          & 78.7                \\ \hline \hline
Ar-CI                   & 43.9                & 43.6 & 45.6          & 45.9          & \textbf{50.5}       \\ \hline
Ar-Pr                   & 62.6                & 57.0 & 59.3          & 61.2          & \textbf{68.0}       \\ \hline
Ar-Rw                   & 68.8                & 67.9 & 70.1          & 68.9          & \textbf{73.6}       \\ \hline
CI-Ar                   & 44.7                & 45.8 & 47.0          & 50.4          & \textbf{51.5}       \\ \hline
CI-Pr                   & 59.5                & 56.5 & 58.5          & 59.7          & \textbf{63.6}       \\ \hline
CI-Rw                   & 59.8                & 60.4 & 60.9          & 61.0          & \textbf{64.8}       \\ \hline
Pr-Ar                   & 46.8                & 44.0 & 46.1          & 45.8          & \textbf{51.5}       \\ \hline
Pr-CO                   & 41.3                & 43.6 & 43.7          & 43.4          & \textbf{45.9}       \\ \hline
Pr-Rw                   & 70.4                & 67.7 & 68.5          & 70.3          & \textbf{73.1}       \\ \hline
Rw-Ar                   & 62.6                & 63.1 & 63.2          & 63.9          & \textbf{65.0}       \\ \hline
Rw-CI                   & 47.9                & 51.5 & 51.8          & \textbf{52.4} & 51.0                \\ \hline
Rw-Pr                   & 76.8                & 74.3 & 76.8          & 76.8          & \textbf{77.3}       \\ \hline
Mean                    & 57.1                & 56.3 & 57.6          & 58.3          & \textbf{61.3}       \\ \hline \hline
Mean3 &65.1	&68.9	&70.6	&\textbf{71.5}	&69.8 \\ \hline
\end{tabular}
\end{table}

Apart from the results of $\widetilde{\mathcal{M}}_x$ and $\mathcal{M}_x$, we also show the results of $\mathcal{M}_z$ on some existing feature-based transfer methods.
This is to show the effectiveness of RTM$_{lr}^d$ as a new independent transfer method.
For subspace learning case, we use JDA, JGSA, and MMIT, and follow the same experimental configuration in Section \ref{S5-2-1}.
For deep feature learning case, we adopt DAN \cite{long2015learning}, DANN \cite{ganin2016domain}, and JAN \cite{tzeng2017adversarial}.
We use the results on ResNet-50 features reported in the corresponding works.
The results are shown in Tables \ref{subsapce} and \ref{deepsupp}.
For each task, the best result is highlighted using bold.

From Table \ref{subsapce}, it can be seen that $\widetilde{\mathcal{M}}_x$ consistently achieves better transfer performance than $\mathcal{M}_x$, and yields 11.58\% improvement in average.
This again verifies the effectiveness of RTM$_{lr}^d$, not only on the new data, but also on the original data.
We further observe that $\widetilde{\mathcal{M}}_x$ also outperforms $\mathcal{M}_z$ in terms of the average performance.
Even as a new independent transfer method, RTM$_{lr}^d$ shows its promising transfer capability, not to say it can further improve the transfer performance based on the existing subspace-based transfer methods.
Table \ref{deepsupp} shows the results of deep learning case.
Not surprisingly, $\widetilde{\mathcal{M}}_x$ beats $\mathcal{M}_x$ in all tasks.
However, different from the subspace case where $\widetilde{\mathcal{M}}_x$ also outperforms $\mathcal{M}_z$, $\widetilde{\mathcal{M}}_x$ is generally not as good as $\mathcal{M}_z$ in the deep learning case.
However, we believe that this is not exactly a negative finding.
On the one hand, we observe $\widetilde{\mathcal{M}}_x$ achieves better results than $\mathcal{M}_z$ in some tasks, e.g., DAN in $\mathbf{A}$-$\mathbf{D}$, $\mathbf{W}$-$\mathbf{A}$, and $\mathbf{P}$-$\mathbf{I}$.
For Office-Home dataset, $\widetilde{\mathcal{M}}_x$ even achieves the best among all the baselines in 11 out of 12 transfer tasks.
On the other hand, $\widetilde{\mathcal{M}}_x$ has a closed-form solution and is computationally efficient.
It only takes minutes to complete learning with moderate computing resources, while deep learning methods usually take hours and even days with tremendous computing powers.
Furthermore, it can be expected that combining RTM$_{lr}^d$ with these deep learning methods can further boost the transfer performance.

\section{Conclusions}
In this paper, we propose a new transfer model, Randomized Transferable Machine (RTM).
We notice that the new feature representation learned by existing feature-based transfer methods is not truly \emph{domain-invariant}, and directly training a transfer model on it is problematic.
To overcome this issue, we propose to learn a transfer model that performs well on infinite source randomly corrupted data populations.
To do so, we propose a marginalized solution to RTM and instantiate an RTM$_{lr}^d$ and an RTM$_{kr}^d$.
Extensive experiments show that our proposed RTM is a promising transfer model.

\ifCLASSOPTIONcompsoc
  \section*{Acknowledgments}
\else
  \section*{Acknowledgment}
\fi
The work was partially done when the first author worked at National University of Singapore. The authors would like to thank Dr. Leong Tze Yun's valuable comments on the work.

\bibliography{RTM}

\begin{thebibliography}{10}
\providecommand{\url}[1]{#1}
\csname url@samestyle\endcsname
\providecommand{\newblock}{\relax}
\providecommand{\bibinfo}[2]{#2}
\providecommand{\BIBentrySTDinterwordspacing}{\spaceskip=0pt\relax}
\providecommand{\BIBentryALTinterwordstretchfactor}{4}
\providecommand{\BIBentryALTinterwordspacing}{\spaceskip=\fontdimen2\font plus
\BIBentryALTinterwordstretchfactor\fontdimen3\font minus
  \fontdimen4\font\relax}
\providecommand{\BIBforeignlanguage}[2]{{%
\expandafter\ifx\csname l@#1\endcsname\relax
\typeout{** WARNING: IEEEtran.bst: No hyphenation pattern has been}%
\typeout{** loaded for the language `#1'. Using the pattern for}%
\typeout{** the default language instead.}%
\else
\language=\csname l@#1\endcsname
\fi
#2}}
\providecommand{\BIBdecl}{\relax}
\BIBdecl

\bibitem{tan2018survey}
C.~Tan, F.~Sun, T.~Kong, W.~Zhang, C.~Yang, and C.~Liu, ``A survey on deep
  transfer learning,'' in \emph{International Conference on Artificial Neural
  Networks}.\hskip 1em plus 0.5em minus 0.4em\relax Springer, 2018, pp.
  270--279.

\bibitem{wang2019softly}
D.~Wang, C.~Lu, J.~Wu, H.~Liu, W.~Zhang, F.~Zhuang, and H.~Zhang, ``Softly
  associative transfer learning for cross-domain classification,'' \emph{IEEE
  transactions on cybernetics}, 2019.

\bibitem{li2019multisource}
J.~Li, S.~Qiu, Y.-Y. Shen, C.-L. Liu, and H.~He, ``Multisource transfer
  learning for cross-subject eeg emotion recognition,'' \emph{IEEE transactions
  on cybernetics}, 2019.

\bibitem{wei2021subdomain}
P.~Wei, Y.~Ke, X.~Qu, and T.-Y. Leong, ``Subdomain adaptation with manifolds
  discrepancy alignment,'' \emph{IEEE Transactions on Cybernetics}, 2021.

\bibitem{Wei2016DeepNF}
P.~Wei, Y.~Ke, and C.~K. Goh, ``Deep nonlinear feature coding for unsupervised
  domain adaptation,'' in \emph{IJCAI}, 2016, pp. 2189--2195.

\bibitem{Long2017Deep}
M.~Long, H.~Zhu, J.~Wang, and M.~I. Jordan, ``Deep transfer learning with joint
  adaptation networks,'' in \emph{ICML}, 2017, pp. 2208--2217.

\bibitem{zhang2018collaborative}
W.~Zhang, W.~Ouyang, W.~Li, and D.~Xu, ``Collaborative and adversarial network
  for unsupervised domain adaptation,'' in \emph{CVPR}, 2018, pp. 3801--3809.

\bibitem{pan2011domain}
S.~J. Pan, I.~W. Tsang, J.~T. Kwok, and Q.~Yang, ``Domain adaptation via
  transfer component analysis,'' \emph{Neural Networks and Learning Systems,
  IEEE Transactions on}, vol.~22, no.~2, pp. 199--210, 2011.

\bibitem{long2013TSC}
M.~Long, G.~Ding, J.~Wang, J.~Sun, Y.~Guo, and P.~S. Yu, ``Transfer sparse
  coding for robust image representation,'' in \emph{CVPR}, 2013, pp. 407--414.

\bibitem{sun2016return}
B.~Sun, J.~Feng, and K.~Sa-enko, ``Return of frustratingly easy domain
  adaptation,'' in \emph{AAAI}, 2016, pp. 2058--2065.

\bibitem{Li2018Domain}
W.~Li, Z.~Xu, D.~Xu, D.~Dai, and G.~L. Van, ``Domain generalization and
  adaptation using low rank exemplar svms,'' \emph{IEEE Transactions on Pattern
  Analysis and Machine Intelligence}, vol.~40, no.~5, pp. 1114 -- 1127, 2018.

\bibitem{wei2019knowledge}
P.~Wei and Y.~Ke, ``Knowledge transfer based on multiple manifolds
  assumption,'' in \emph{CIKM}, 2019, pp. 279--287.

\bibitem{cai2019learning}
R.~Cai, Z.~Li, P.~Wei, J.~Qiao, K.~Zhang, and Z.~Hao, ``Learning disentangled
  semantic representation for domain adaptation,'' in \emph{IJCAI}, 2019, pp.
  2060--2066.

\bibitem{borgwardt2006integrating}
K.~M. Borgwardt, A.~Gr-etton, M.~J. Rasch, H.-P. Kriegel, B.~Sch-{\"o}lkopf,
  and A.~J. Smola, ``Integrating structured biological data by kernel maximum
  mean discrepancy,'' \emph{Bioinformatics}, vol.~22, no.~14, pp. 49--57, 2006.

\bibitem{ganin2016domain}
Y.~Ganin, E.~Ustinova, H.~Ajakan, P.~Germain, H.~Larochelle, F.~Laviolette,
  M.~Marchand, and V.~Lempitsky, ``Domain-adversarial training of neural
  networks,'' \emph{The Journal of Machine Learning Research}, vol.~17, no.~1,
  pp. 2096--2030, 2016.

\bibitem{lee2019sliced}
C.-Y. Lee, T.~Batra, M.~H. Baig, and D.~Ulbricht, ``Sliced wasserstein
  discrepancy for unsupervised domain adaptation,'' in \emph{CVPR}, 2019, pp.
  10\,285--10\,295.

\bibitem{wei2021randomized}
P.~Wei and T.~Y. Leong, ``Randomized transferable machine,'' in \emph{2020 25th
  International Conference on Pattern Recognition (ICPR)}.\hskip 1em plus 0.5em
  minus 0.4em\relax IEEE, 2021, pp. 8711--8718.

\bibitem{gong2012geodesic}
B.~Gong, Y.~Shi, F.~Sha, and K.~Grauman, ``Geodesic flow kernel for
  unsupervised domain adaptation,'' in \emph{CVPR}, 2012, pp. 2066--2073.

\bibitem{fernando2013unsupervised}
B.~Fernando, A.~Habrard, M.~Sebban, and T.~Tuytelaars, ``Unsupervised visual
  domain adaptation using subspace alignment,'' in \emph{ICCV}, 2013, pp.
  2960--2967.

\bibitem{long2013transfer}
M.~Long, J.~Wang, G.-g. Ding, J.~Sun, and P.~S. Yu, ``Transfer feature learning
  with joint distribution adaptation,'' in \emph{ICCV}, 2013, pp. 2200--2207.

\bibitem{long2014transfer}
------, ``Transfer joint matching for unsupervised domain adaptation,'' in
  \emph{CVPR}, 2014, pp. 1410--1417.

\bibitem{zhang2017joint}
J.~Zhang, W.~Li, and P.~Ogunbona, ``Joint geometrical and statistical alignment
  for visual domain adaptation,'' in \emph{CVPR}, 2017, pp. 692--701.

\bibitem{chen2012marginalized}
M.~Chen, Z.~Xu, K.~Weinberger, and F.~Sha, ``Marginalized denoising
  autoencoders for domain adaptation,'' in \emph{ICML}, 2012, pp. 1627--1634.

\bibitem{long2015learning}
M.~Long and J.~Wang, ``Learning transferable features with deep adaptation
  networks,'' in \emph{ICML}, 2015, pp. 97--105.

\bibitem{Yan2017Mind}
H.~Yan, Y.~Ding, P.~Li, Q.~Wang, Y.~Xu, and W.~Zuo, ``Mind the class weight
  bias: Weighted maximum mean discrepancy for unsupervised domain adaptation,''
  in \emph{CVPR}, 2017, pp. 945--954.

\bibitem{tzeng2017adversarial}
E.~Tzeng, J.~Hoffman, K.~Saenko, and T.~Darrell, ``Adversarial discriminative
  domain adaptation,'' in \emph{CVPR}, 2017, pp. 7167--7176.

\bibitem{shorten2019survey}
C.~Shorten and T.~M. Khoshgoftaar, ``A survey on image data augmentation for
  deep learning,'' \emph{Journal of Big Data}, vol.~6, no.~1, p.~60, 2019.

\bibitem{Dimitris2019Robust}
\BIBentryALTinterwordspacing
D.~Tsipras, S.~Santurkar, L.~Engstrom, A.~Turner, and A.~Madry, ``Robustness
  may be at odds with accuracy,'' in \emph{ICLR}, 2019. [Online]. Available:
  \url{https://openreview.net/forum?id=SyxAb30cY7}
\BIBentrySTDinterwordspacing

\bibitem{zakharov2019deceptionnet}
S.~Zakharov, W.~Kehl, and S.~Ilic, ``Deceptionnet: Network-driven domain
  randomization,'' \emph{ICCV}, pp. 1627--1634, 2019.

\bibitem{Bishop2006}
C.~M. Bishop, \emph{Pattern Recognition and Machine Learning (Information
  Science and Statistics)}.\hskip 1em plus 0.5em minus 0.4em\relax Secaucus,
  NJ, USA: Springer-Verlag New York, Inc., 2006.

\bibitem{wei2018feature}
P.~Wei, Y.~Ke, and C.~K. Goh, ``Feature analysis of marginalized stacked
  denoising autoenconder for unsupervised domain adaptation,'' \emph{IEEE
  transactions on neural networks and learning systems}, pp. 1--14, 2018.

\bibitem{wager2013dropout}
S.~Wager, S.~Wang, and P.~S. Liang, ``Dropout training as adaptive
  regularization,'' in \emph{Advances in Neural Information Processing
  Systems}, 2013, pp. 351--359.

\bibitem{li2017domain}
W.~Li, Z.~Xu, D.~Xu, D.~Dai, and L.~Van~Gool, ``Domain generalization and
  adaptation using low rank exemplar svms,'' \emph{IEEE transactions on pattern
  analysis and machine intelligence}, vol.~40, no.~5, pp. 1114--1127, 2017.

\bibitem{clinchant2016domain}
S.~Clinchant, G.~Csurka, and B.~Chidlovskii, ``A domain adaptation
  regularization for denoising autoencoders,'' in \emph{ACL}, 2016, pp. 26--31.

\bibitem{blitzer2007biographies}
J.~Blitzer, M.~Dredze, F.~Pereira \emph{et~al.}, ``Biographies, bollywood,
  boom-boxes and blenders: Domain adaptation for sentiment classification,'' in
  \emph{ACL}, vol.~7, 2007, pp. 440--447.

\bibitem{jarvelin2002cumulated}
K.~J\"{a}rvelin and J.~Kek\"{a}l\"{a}inen, ``Cumulated gain-based evaluation of
  ir techniques,'' \emph{ACM Trans. Inf. Syst.}, vol.~20, no.~4, pp. 422--446,
  Oct. 2002.

\end{thebibliography}
\bibliographystyle{IEEEtran}

\end{document}